**Machine Learning For An Explainable Cost Prediction of Medical Insurance**

Ugochukwu Orji[1] and Elochukwu Ukwandu[2,*]

[1]Department of Computer Science, Faculty of Physical Science, University of Nigeria, Nsukka, Enugu State, Nigeria.
[2]Department of Applied Computing and Engineering, Cardiff School of Technologies, Cardiff Metropolitan University, Cardiff CF5 2YB, Wales, United Kingdom.

**Email addresses**: ugochukwu.orji.pg00609@unn.edu.ng, eaukwandu@cardiffmet.ac.uk
[*]**Correspondent Author**: Elochukwu Ukwandu (eaukwandu@cardiffmet.ac.uk)

**Abstract**
Predictive modeling in healthcare continues to be an active actuarial research topic as more insurance companies aim to maximize the potential of Machine Learning (ML) approaches to increase their productivity and efficiency. In this paper, the authors deployed three regression-based ensemble ML models that combine variations of decision trees through Extreme Gradient Boosting (XGBoost), Gradient-boosting Machine (GBM), and Random Forest (RF) methods in predicting medical insurance costs. Explainable Artificial Intelligence (XAi) methods SHapley Additive exPlanations (SHAP) and Individual Conditional Expectation (ICE) plots were deployed to discover and explain the key determinant factors that influence medical insurance premium prices in the dataset. The dataset used comprised 986 records and is publicly available in the KAGGLE repository. The models were evaluated using four performance evaluation metrics, including R-squared ($R^2$), Mean Absolute Error (MAE), Root Mean Squared Error (RMSE), and Mean Absolute Percentage Error (MAPE). The results show that all models produced impressive outcomes; however, the XGBoost model achieved a better overall performance although it also expanded more computational resources, while the RF model recorded a lesser prediction error and consumed far fewer computing resources than the XGBoost model. Furthermore, we compared the outcome of both XAi methods in identifying the key determinant features that influenced the PremiumPrices for each model and whereas both XAi methods produced similar outcomes, we found that the ICE plots showed in more detail the interactions between each variable than the SHAP analysis which seemed to be more high-level. It is the aim of the authors that the contributions of this study will help policymakers, insurers, and potential medical insurance buyers in their decision-making process for selecting the right policies that meet their specific needs.

# 1 Introduction

In recent times, actuarial modeling of insurance claims has become a key research area in the health insurance sector and is mostly applied in setting effective premiums (Duncan *et al.*, 2016). This is vital for attracting and retaining insureds and for efficient management of existing plan members. However, due to the variety of factors that drive medical insurance costs and the complexities therein, there is a bit of a challenge being faced in accurately building a predictive model for it. Factors like demographic information, health status, geographic access, lifestyle choices, provider characteristics, etc. all have the potential to have a dramatic impact on the expected costs of medical insurance. Other vital factors like the scope of coverage, type of plan, deductible, and the age of a customer when they sign up also play a major role in determining the potential cost of medical insurance.

The importance of an effective and transparent medical insurance system cannot be overemphasized considering the need for universal healthcare coverage and the challenges thrown up by the COVID-19 pandemic (Orji *et al.*, 2022b).

Orji *et al.*, (2022a) described how the predictive analytics feature of ML has become the most utilized feature for industrial applications. The ongoing regulatory and market changes in the health industry continue to motivate actuarial research into predictive modeling in health care. ML algorithms have proven to yield accurate results in predicting high-cost, high-need patient expenditures, so insurance companies are increasingly turning to ML approaches to improve their policies and premium settings (Yang *et al.*, 2018).

However, the high performance of ML algorithms in healthcare, as noted by (Panay *et al.*, 2019), is somewhat offset by their black-box nature. If the predictive analytics involving patients' personal and clinical information is not well understood or explained, they could be prone to bias. Fortunately, the emergence of Explainable Artificial Intelligence (XAi) methods allows those involved - patients, healthcare managers, and insurers - to gain further insight into the logic behind predictions, consequently promoting transparency and acceptability. As a result, predicting medical insurance costs with high certainty would solve problems such as accountability and transparency, enabling control over all parties involved in patient care (Panay *et al.*, 2019).

## 1.1 Research Significance

The contributions of this study are as follows:

(i) Predicting medical insurance costs using ensemble learning-based machine learning models (XGBoost, GBM, and RF).
(ii) Comparing the performance outcome of the algorithms deployed via various evaluation metrics.
(iii) Explaining the BlackBox models using XAi methods and interpreting the determinants that influence prediction outcomes.
(iv) Comparing the outcome of both XAi methods in explaining the black-box models

The rest of the paper is structured as follows: the concept of medical insurance and related works are discussed in Section 2. Section 3 describes the methodology and approaches taken. Section 4 contains the experimental outcomes and the result of the determinant analysis. Finally, Section 5 summarizes the findings and provides recommendations for future work.

# 2  Literature Review

## 2.1  Concept of Medical Insurance and Its Benefits

Generally, health insurance is a pre-payment and risk-pooling mechanism designed to cover medical expenses incurred due to an illness. These expenses include hospitalizations, medicines, and doctor's visits. Social and national health insurance has now made it possible to improve equitable access to health care and protect people from financial risks associated with diseases (ISSA, 2021).

For most health insurance systems available today, there are two main financing arrangements:

1. The single health insurance pool (single-payer)
2. The multiple health insurance pool (multiple-payer)

Also, within both financing arrangements, there exist many variations. The single-payer option generally includes greater government control over the provision of care and tends to emphasize equity. While the multi-payer option allows consumers to choose from the insurance companies available and this approach has been shown to drive innovation and competition (ISSA, 2021). In many countries, a combination of both single and multi-payer insurance is obtainable.

### 2.1.1  Major types of health insurance systems

There are about three major types of health insurance systems representing specific combinations:

- Health Maintenance Organisation (HMO): In this plan, a list of doctors working directly with the HMO or on contract is provided for the insureds to pick as primary care physicians.
- Preferred Provider Organization (PPO): PPO plans work by providing a list of pre-approved contracted providers. The use of a 60/40 split for reimbursement is common, which means the insurer pays 60 percent of the costs, and the insured covers the remaining 40 percent.
- High-Deductible Health Plan (HDHP) With a Health Savings Account (HSA): HDHPs work by getting the insured to operate a health savings account with the provider from where treatment costs are deducted on an agreed percentage and plan with likely lower monthly premiums.

Additionally, most medical insurance companies compensate customers for their medical expenses in two ways (Maheshwari, 2023) and (Kagan, 2023):

- Cashless Treatment: Here, the insurance company pays the hospital directly rather than the policyholder upfront.
- Reimbursement: It is expected that the policyholder will settle his or her medical expenses with the hospital before seeking reimbursement from the insurance company.

### 2.1.2  Benefits of Medical Insurance Coverage

The following are some benefits of medical insurance for customers and healthcare managers:

- *Coverage against critical illness* - some diseases can be prolonged and fatal in nature, like renal failure, heart ailments, and cancer. For these types of illnesses, their plans operate differently with regard to making payments for benefits. However, with the right insurance plan these illnesses will be covered for the customer, thus guaranteeing the needed care is received.
- *Safety against rising medical costs* - just like with most economic sectors, the healthcare industry is experiencing an economic downturn with incessant rising costs. Directly paying money for healthcare can be extremely expensive. Under these circumstances, health insurance has the advantage. When an individual considers the medical insurance policy in a financial plan, it can reduce the trouble. The safety assurance against rising healthcare charges is a remarkable benefit of medical insurance.
- *Insurance coverage for healthcare emergencies* - health emergencies are unpredictable and can occur at random. With a good medical insurance policy, one need not worry about the expenses, but rather concentrate and full recovery from the illness.
- *Improvements in medical technology and capacity* - *a*ccording to a study by Weisbrod (1991), the expansion of health insurance coverage over the years has created incentives to develop new medical technologies. This was corroborated by (Clemens & Olsen, 2021), who demonstrated that 20 to 30 percent of medical equipment and device innovation between 1965 and 1990 could be attributed to Medicare/Medicaid's introduction in 1965. Furthermore, Frankovic & Kuhn (2023) conducted an expansive macroeconomic analysis of how health insurance influenced medical progress in the US from 1965 to 2005, considering its direct impact on lowering out-of-pocket costs as well. The results showed that insurance expansion accounts for approximately 63 percent of the growth in healthcare expenditure between 1965 and 2005, further stimulating medical R&D with an additional 57 percent.

### 2.1.3 Challenges of Medical Insurance

Undeniably, all countries face major challenges in maintaining a sustainable healthcare system. Depending on the country, these challenges include issues such as the impact of aging populations, the rising chronic disease burden, health infrastructure development, and generally the financing of healthcare services.

Using the US as a case study, the US per capita health expenditure was $7,960 in 2009 (Publisher, 2016) but according to Statista, the 2021 figures stood at $10,202 per resident (Statista, 2022). However, compared to other countries in the Organisation for Economic Co-operation and Development (OECD), the expenditure by the United States seems huge especially as the United States lags behind many of its peer nations in several indicators of health and health care quality (Nolte & McKee, 2008).

So why is the US healthcare spending much higher? Research has traced this to two major issues namely:

i. The United States has the highest administrative costs for health care in the industrial world, estimated at around $360 billion a year, or 14 percent of all US healthcare expenses (Emmanuel, 2011). These costs are mainly due to the dominance of private insurance which necessitates heavy clerical and accounting demands that could have been more effectively managed with the implementation of electronic technologies (Boffey, 2012).

ii. Second, private insurance in the United States is based on a fee-for-service model. Physicians, hospitals, healthcare professionals, and businesses can charge whatever they want for their services under this model. Compared to other industrial nations with lower prices, the United States has a fundamentally different healthcare system from its peer nations which helps explain why healthcare services are so much more expensive in the United States than in its peer nations (Klein, 2012).

Conclusively, ML-driven health insurance has the capacity to bring about better outcomes and tackle some of the most formidable complications within the healthcare system, especially in America. As demands on consumers and providers increase and the sector struggles with a deficit of simplicity and technology, ML can provide a cheaper, more efficient solution. By gaining an insight into individual issues and specific healthcare needs, AI and ML-based approaches can organize data, while simultaneously finding innovative ways to simplify the industry along with improving patient safety and results.

## 2.2 Machine Learning: A Brief Introduction

Machine learning (ML) is a subset of Artificial intelligence (AI) that focuses on creating intelligent systems capable of autonomous learning from vast datasets. As emphasized by various studies (Carleo et al., 2019; Akter, 2020) while AI encompasses various technologies, including expert systems, deep learning, and robotics, ML specifically revolves around data-driven learning.

According to authors (Adibimanesh *et al.,* 2023; Fernández & Peters, 2023) the widespread adoption of ML techniques across various industries, can be attributed to advancements in ML techniques, the computational power of modern GPUs, and the availability of diverse datasets. Despite these achievements, it is noted that the full potential of AI and ML has yet to be realized, and there is ongoing research in this area (Panesar, 2019).

ML tools and techniques help in decision-making through prediction and forecasting based on data. Generally, the more data an ML algorithm has, the better it performs (Ngiam & Khor, 2019), especially in medical-related applications, as described by (Ray, 2019).

**The three categories of ML**

1. **Supervised Learning:** supervised learning maps the association between inputs and outputs of a set of labeled training data (Qayyum, 2020). To demonstrate this phenomenon: Imagine an n set of sample data $\{X_\mu, y_\mu\}_{\mu=1,\ldots,n}$ we can denote one sample of the data as $X_\mu \in \mathbb{R}^p$ with $\mu = 1, \ldots, n$ where each $X_\mu$ can be an image, and $P$ the number of pixels available in the image. Also, each $X_\mu$ sample contains a label $y_\mu \in \mathbb{R}^d$, where $d = 1$. The label could contain an attribute of the image. The supervised learning technique tries to identify a function *f*, which, on receiving an input $X_{new}$ without a label, can accurately predict its output *f*($X_{new}$). To evaluate the performance of this function *f*, data samples available are typically separated into two sets - the training set used for developing the function and the test set for measuring how well it works. Linear and logistic regression, support vector machines (SVM), as well as ensemble approaches like random forest and XGBoost, are some of the famous methods employed in such tasks.

2. **Unsupervised Learning:** Unsupervised Machine Learning, unlike supervised ML, lacks predefined labels. It autonomously discovers patterns within data, making it effective for identifying hidden trends. Common applications include clustering with algorithms like K-Means and DBSCAN.
3. **Semi-supervised Learning:** Semi-supervised Learning (SSL) combines labeled and unlabeled data for classification or clustering tasks (Van Engelen & Hoos, 2020). The authors (Zhu & Goldberg, 2009) identified SSL's main category to include inductive SSL (predicting labels for unseen data) and transductive SSL (estimating labels for current unlabeled samples). SSL is applied in Speech Analysis, Protein sequence classification, and Text document classification.

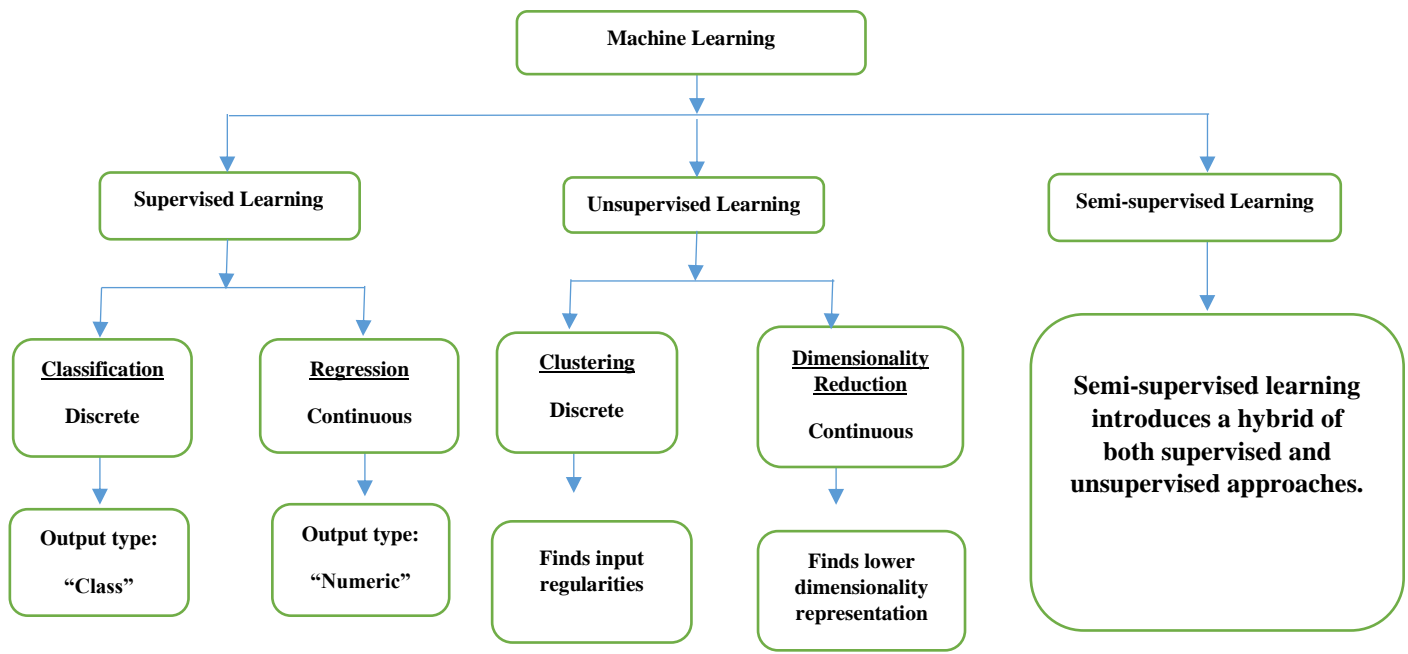

*Figure 1: Overview of ML Categories (Ikerionwu, et al., 2022)*

## 2.3 Different ML algorithms used in related works

Over the last ten years, health insurers have increasingly turned to AI and ML to identify people in need of greater protection and improve the insurance underwriting process. A system aiding consumers in purchasing health coverage must process data including hospital engagements, the ages of insureds, financial risk comfort level, degree of security, and many other variables. While ML is recognized for its data-intensive purposes, its biggest strength when handling a customer's healthcare comes down to its capacity for proficient deduction and quick reading of trends (Pomales, 2022).

Industry leaders are realizing that ML applications can help improve the accuracy of treatment protocols and health outcomes as more healthcare and medical companies adopt AI and ML in their varied systems. It's essential to understand that the greater part of what is charged when buying health insurance goes into risk prediction and risk management and other overhead expenditures. By employing AI to build a system with more exact risk models, thereby determining which customers require specialized care, health insurers can allocate larger amounts towards their beneficiaries rather than these activities. Platforms incorporating data, refining evaluations, and producing intelligent observations can also significantly reduce the need for costly human analysts and other administrative costs.

Much research has been done on using ML systems to forecast medical insurance costs, as indicated in (ul Hassan, *et al*., 2021), where Linear Regression, Support Vector Regression, Ridge Regressor, Stochastic Gradient Boosting (SGB), XGBoost, Decision Tree, Random Forest Regressor, Multiple Linear Regression, and k-Nearest Neighbors were used. This study employed the medical insurance dataset obtained from KAGGLE. In the end, SGB had the highest accuracy at 86% with an RMSE of 0.340.

Medical insurance research has more recently focused on developing models to accurately predict High-cost Claimants (HiCCs). These people account for 9% of all healthcare costs in the US, as their annual costs exceed $250,000. To this end, Maisog *et al.* (2019) conducted a study with 48 million individuals that incorporated health insurance claims and census data. Various ML models such as Random Forests, Support Vector Machines, Gradient Boosted Trees, Elastic Nets, and XGBoost were employed to calculate the individual risk of being a HiCC. Among them, LightGBM held the highest AUC-ROC score at 91.25%, reiterating that claims data can produce high-performing predictive models.

Duncan *et al.* (2016) utilized a dataset comprising 30,000 insureds from the USA that were received from health actuarial consultants Solucia Inc. (now SCIO Health Analytics) to construct ML models like the Linear model, Lasso, and Elastic Nets, Multivariate Adaptive Regression Splines (MARS), Random Forest, M5 Decision Trees, and Generalized Boosting Models to forecast future healthcare costs. The outcomes reflected that ML approaches are more appropriate for use in healthcare than the customary linear methods. M5, RF, and GBM emerged as the best-performing models among the Truncated $R^2$, Truncated MAE, and $qMAD_{0.95}$ metrics implemented.

The authors (Taloba *et al*., 2022) employed a business analytic method that incorporated statistical and ML approaches to determine the health costs resulting from obesity via linear regression analysis, Naive Bayes classifier, and Random Forest algorithms. Using a dataset of 24,353 anonymized patients from Tsuyama Chuo Hospital, the results pointed to the maximum accuracy of 97.89 percent in forecasting total healthcare costs via the Linear Regression model.

Panay *et al.* (2019) introduced Interpretable Evidence Regression (IEVREG), an extended version of the evidence regression model (EVREG), for the purpose of predicting healthcare costs utilizing electronic health records consisting of 25,464 Japanese health records from Tsuyama Chuo Hospital's 2016 to 2017 medical checkups, exam results, and billing information. IEVREG achieved the highest performance with an $R^2 = 0.44$ compared to the Artificial Neural Network and Gradient Boosting method with $R^2$ of 0.40 and 0.35 respectively.

Finally, the authors of a related study, Vimont *et al.* (2022), compared a simple neural network (NN) and Random Forest (RF) to a generalized linear model (GLM) for predicting medical expenses at the individual level. The dataset used is a 1/97 representative sample of the French National Health Data Information System, which contains about 510,182 patient records. The result showed that the RF model obtained the best performance reaching an adj- $R^2$ of 47.5%, an MAE of 1338€, and a hit ratio (HiR) of 67%.

## 2.4 Use of XAi in Related Works

Indeed, XAi methods have in recent times been deployed by researchers in various research areas including medical insurance cost prediction. Below are some surveyed literature:

The authors in Kshirsagar *et al.* (2021) assessed the effectiveness of ML models in predicting per-member per-month costs for employer groups during health insurance renewals, particularly focusing on groups with cost-saving potential. The research involved the development of two models, one at the individual patient level and another at the employer-group level, to estimate the annual per-member per-month allowed amount for employer groups, using a dataset of 14 million patients. The results showed a 20% improvement in performance compared to the existing insurance carrier's pricing model and successfully identified 84% of the concession opportunities. This underscores the efficacy of ML systems in establishing accurate and equitable health insurance pricing. Additionally, the SHAP XAi method was applied to the LightGBM model to provide member-level explanations for rate adjustments, enhancing the model's robustness.

Bora *et al.*, (2022) proposed an ML-based health insurance price prediction via two regression models; multi-linear regression and random forest. They further deployed XAi methods namely Microsoft InterpretML, Local Interpretable Model-agnostic Explanations (LIME), and SHAP to explain their black-box models. The study was able to establish that XAi methods are effective in explaining black-box models for medical insurance prediction and the effectiveness of XAi in making predictions more reliable and transparent.

Also, Langenberger *et al.*, (2023) in their research, used healthcare claims data to predict future high-cost patients via various ML methods like Random Forest (RF), Gradient Boosting Machine (GBM), Artificial Neural Network (ANN), and Logistic Regression (LR). The result of the study showed that the RF and GBM models achieved the best performances with the RF model producing optimal results. Furthermore, the authors only applied the XAi method SHAP on the RF model which was the best-performing model in their study. Through the SHAP analysis, they were able to explore how the variables influenced predicted class probabilities in the study.

In another related work by (Sahai *et al.*, 2022), a comparative analysis between tree-based classifiers such as Decision Tree, Random Forest, and XGBoost was carried out. The study focused on enhancing risk assessment capabilities for life insurance companies using predictive analytics by classifying the insurance risks based on historical data. The findings revealed that XGBoost outperformed the others. Additionally, the authors underscored the significance of model interpretability, particularly for stakeholders within the insurance sector.

Finally, in this study, we deployed two XAi methods; SHAP and ICE for a deeper analysis of how individual features in the dataset affect the overall outcome of our models. We further compared the performance of each XAi method deployed. To the best of our knowledge, none of the previous studies in this field have performed such analysis; thus, our work contributes fresh perspective to the body of knowledge in the field.

*Table 1: Summary of Reviewed Works*

| Ref. | Forecasting method | Data source | Performance metrics | Outcome/Significance |
|---|---|---|---|---|
| ul Hassan *et al.* (2021) | Linear Regression, Support Vector Regression, Ridge Regressor, Stochastic Gradient Boosting, XGBoost, Decision Tree, Random Forest Regressor, Multiple Linear Regression, and k-Nearest Neighbours. | Medical insurance cost dataset from the KAGGLE repository. About 70% of the total data was used for training while the remaining 30% test data was used for evaluation. | SGB had the highest accuracy at 86% with an RMSE of 0.340. | Analyzing the performance of popular machine learning algorithms for forecasting healthcare costs |
| Maisog *et al.* (2019) | Random Forests, Support Vector Machines, Gradient Boosted Trees, Elastic Nets, Extreme Gradient Boosting, and ensembles | Health insurance claims from 48 million people augmented with census data. Due to size limitations, a subpopulation of 3 million members was used for the training dataset by down sampling the majority class while the final model was evaluated using a dataset of 9,684,279 members. | AUC-ROC of 91.25% was achieved by the Light Gradient Boosted Trees Classifier (LightGBM) algorithm. | Even for rare high-cost claims exceeding $250,000, high-performing predictive models can be constructed using claims data. |
| Duncan *et al.* (2016) | Linear model, Lasso, and Elastic Nets, Multivariate Adaptive Regression Splines (MARS), Random Forest (RF), M5 Decision Trees, and Generalized Boosting Models (GBM). | 30,000 insureds from across the United States are included in the dataset, which was compiled by the health actuarial consultants Solucia Inc. (now SCIO Health Analytics). 2/3 (N = 20,000) of the data was randomly used for training while 1/3 | M5, RF, and GBM were the top-performing models across the Trunc $R^2$, Trunc MAE, and qMAD$_{0.95}$ metrics used. | Comparing ML approaches to standard linear approaches, the findings indicate that ML approaches are much better suited to the healthcare context. |

| | | (M = 10,000) of the data was used for testing. | | |
|---|---|---|---|---|
| Taloba *et al.* (2022) | Linear regression, naive Bayes classifier, and random forest algorithms were compared using a business analytic method that applied statistical and machine learning approaches. | Tsuyama Chuo Hospital provided anonymized medical expense data for 24,353 patients in the dataset used. About 75% of the dataset was used for training and 25% for testing. | In terms of forecasting overall healthcare costs, linear regression had a maximum accuracy of 97.89 percent. | A major objective of this study is to estimate the health costs associated with obesity in the population. |
| Panay *et al.* (2019) | Interpretable Evidence Regression (IEVREG), an extended version of the evidence regression model (EVREG), Artificial Neural Network, and Gradient Boosting method. | 25,464 Japanese health records from Tsuyama Chuo Hospital's 2016 to 2017 medical checkups, exam results, and billing information. About 70% of the data was used for training and 30% for evaluation. | IEVREG achieved the highest performance with an $R^2 = 0.44$ compared to Artificial Neural Network and Gradient Boosting method with $R^2$ of 0.40 and 0.35 respectively. | A new regression method for predicting healthcare costs was presented |
| Vimont *et al.* (2022) | Neural network (NN), random forest (RF), and a generalized linear model (GLM) | The dataset used consists of a 1/97 representative sample of the French National Health Data Information System totaling about 510,182 patient records. Train/test data split not specified. | The result showed that the RF model obtained the best performance reaching an adj-$R^2$ of 47.5%, an MAE of 1338€, and a hit ratio (HiR) of 67%. | Compared a simple neural network (NN) and random forest (RF) to a generalized linear model (GLM) in predicting medical costs. Concluded that the RF model was preferred when the objective is to best predict medical costs while GLM was preferred when the objective is to understand the contribution of predictors. |

# 3 Methodology
## 3.1 Experiment setup
Python libraries on Jupyter Notebook running on an Intel(R) Iris(R) XE GPU, a 2.80 GHz Intel(R) Core (TM) i7-1165G7 CPU, and 16GB RAM was used to perform the experiments. The steps of our working methodology are shown in Figure 2 below.

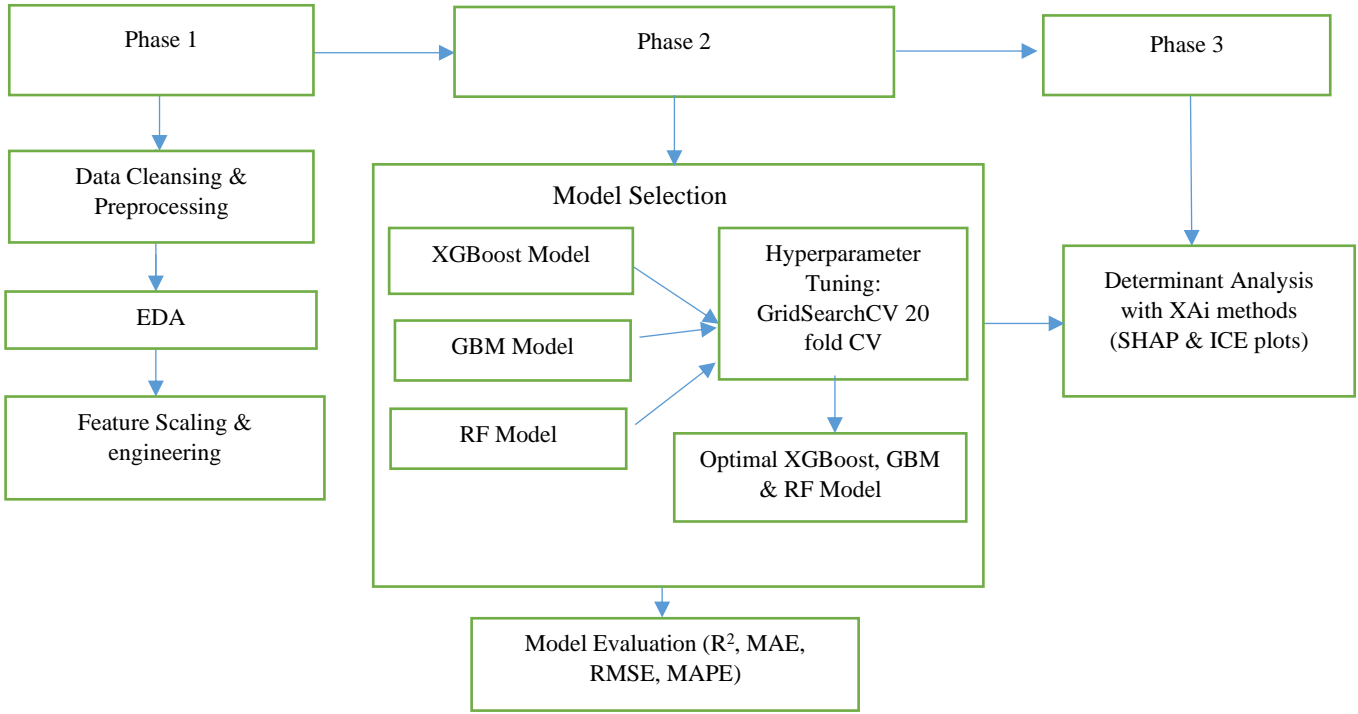

*Figure 2: Workflow of the Experiment*

## 3.2 Data source and preprocessing
The medical insurance cost dataset used was sourced from KAGGLE's repository (Kaggle, 2021). The dataset consists of 986 records and 11 attributes/features as described in Table 2. Thereafter, data preprocessing was done, which is an essential part of any knowledge discovery from data-based projects. Checks for missing and duplicate values were done and no duplicate was found. The data was further normalized using Standard scalar technique.

*Table 2: Dataset Description*

| Type | Name | Description | Data type |
|---|---|---|---|
| Input Variables | Age | Age Of Customer | Numeric |
| | Diabetes | Diabetes status of customer (diabetic/non-diabetic) | Numeric |
| | BloodPressureProblems | Customer blood pressure status | Numeric |
| | AnyTransplants | Any Major Organ Transplants | Numeric |
| | AnyChronicDiseases | Chronic Ailments include; asthma, heart disease, etc. | Numeric |

|  | Height | Height Of Customer | Numeric |
|---|---|---|---|
|  | Weight | Weight Of Customer | Numeric |
|  | KnownAllergies | Any Known Allergies | Numeric |
|  | HistoryOfCancerInFamily | Checks if Any Blood Relative Of The Customer Has Had Any Form Of Cancer | Numeric |
|  | NumberOfMajorSurgeries | The Number Of Major Surgeries That The customer Has Had | Numeric |
| Output Variable | Premium Price | Insurance Premium Price | Numeric |

## 3.3 Exploratory Data Analysis (EDA)

EDA was performed to quickly explore the dataset in a bid to uncover any implicit patterns in the dataset, spot anomalies, test hypotheses, and check assumptions. The information gained from EDA help researchers choose appropriate ML approaches to solving the needed problem (Patil, 2018).

At this stage of the research, it is vital to check the relationships between some key features to identify their correlation as shown by the Pearson correlation Heatmap in figure 3. The Correlation Heatmap shows that there is no strong correlation between the various independent variables aside from the age attribute. This is not uncommon according to the World Health Organization (WHO), increasing age is frequently associated with increased health-care utilization and costs (Universal Health Coverage and Ageing, 2023). Furthermore, Figure 4a and 4b gives an overview of the relationship between various features and the premium price and it shows that Premium Price surges for patients above 30 years. From the weight and height values of each record, the authors calculated the Body Mass Index (BMI) as follows: weight (kg) / [height (m)] $^2$.

Table 3 shows the Statistical summary of the features, it can be observed that the data shows a normal distribution.

*Table 3: Statistical Summary of the Features*

| Features | Mean | STD. | Min. | Q1 (25%) | Median (50%) | Q3 (75%) | Max. |
|---|---|---|---|---|---|---|---|
| Age | 41.75 | 13.96 | 18.00 | 30.00 | 42.00 | 53.00 | 66.00 |
| Diabetes | 0.42 | 0.49 | 0.00 | 0.00 | 0.00 | 1.00 | 1.00 |
| BloodPressureProblems | 0.47 | 0.50 | 0.00 | 0.00 | 0.00 | 1.00 | 1.00 |
| AnyTransplants | 0.06 | 0.23 | 0.00 | 0.00 | 0.00 | 0.00 | 1.00 |
| AnyChronicDiseases | 0.18 | 0.38 | 0.00 | 0.00 | 0.00 | 0.00 | 1.00 |
| Height | 168.18 | 10.10 | 145.00 | 161.00 | 168.00 | 176.00 | 188.0 |
| Weight | 76.95 | 14.27 | 51.00 | 67.00 | 75.00 | 87.00 | 132.00 |
| KnownAllergies | 0.22 | 0.41 | 0.00 | 0.00 | 0.00 | 0.00 | 1.00 |
| HistoryOfCancerInFamily | 0.12 | 0.32 | 0.00 | 0.00 | 0.00 | 0.00 | 1.00 |
| NumberOfMajorSurgeries | 0.67 | 0.75 | 0.00 | 0.00 | 1.00 | 1.00 | 3.00 |
| Premium Price | 24336.71 | 6248.18 | 15000.00 | 21000.00 | 23000.00 | 28000.00 | 40000.00 |

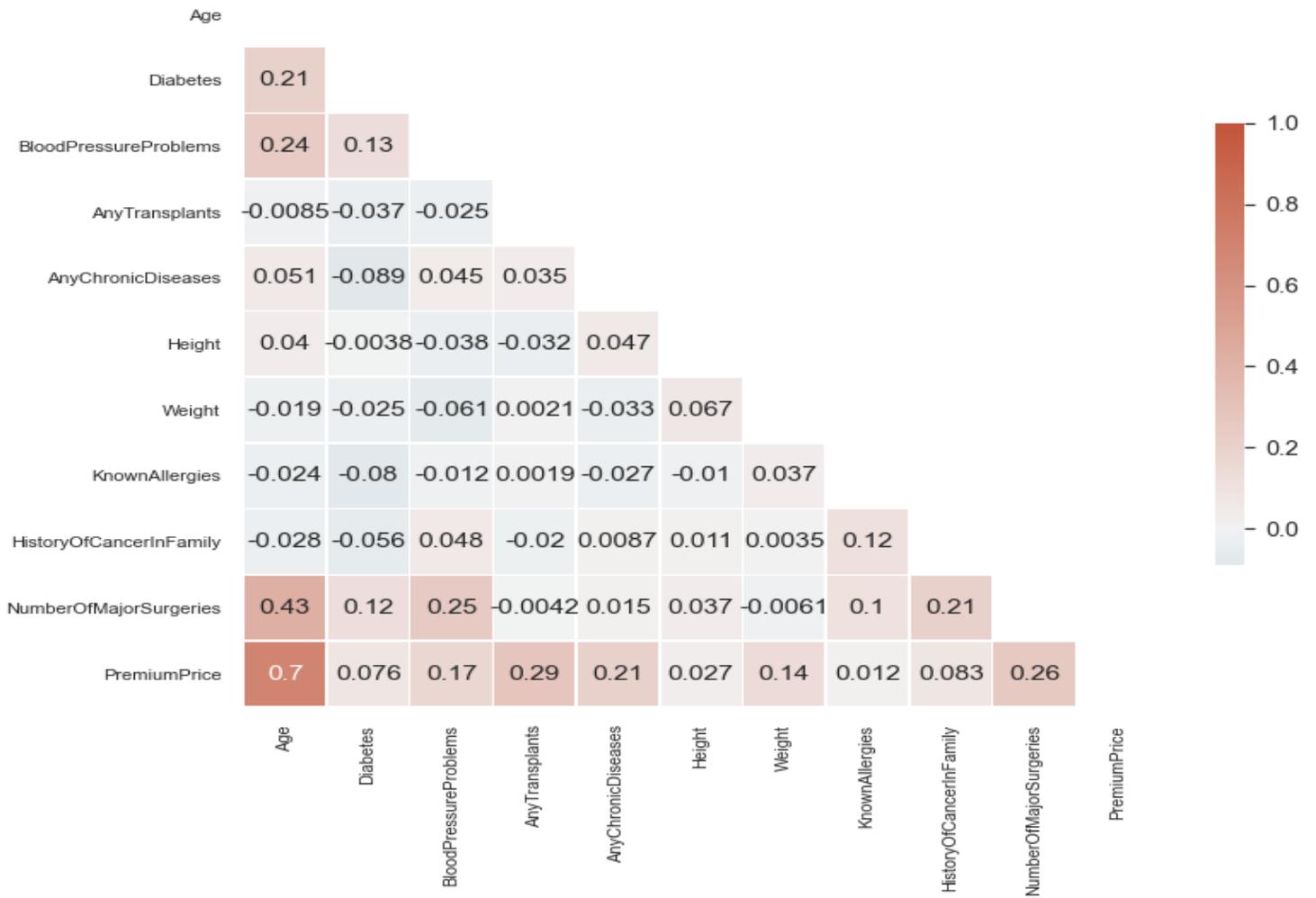

*Figure 3: Correlation Heatmap*

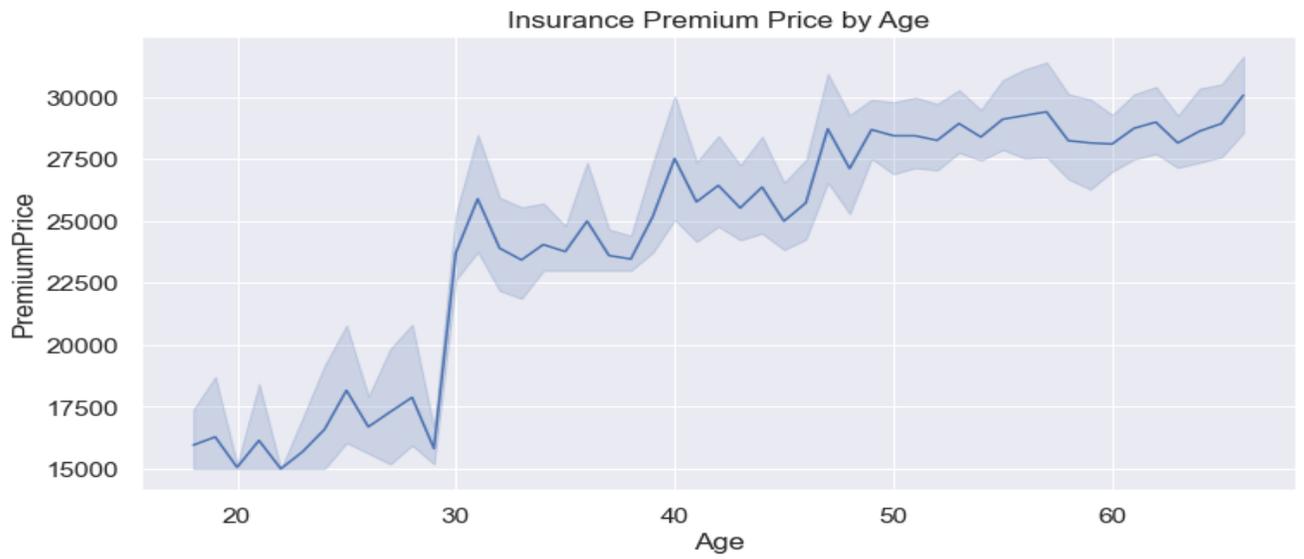
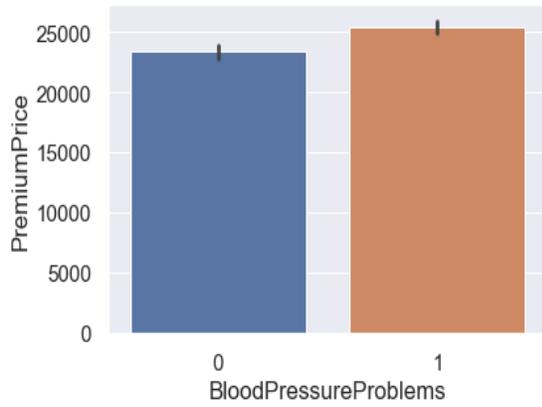
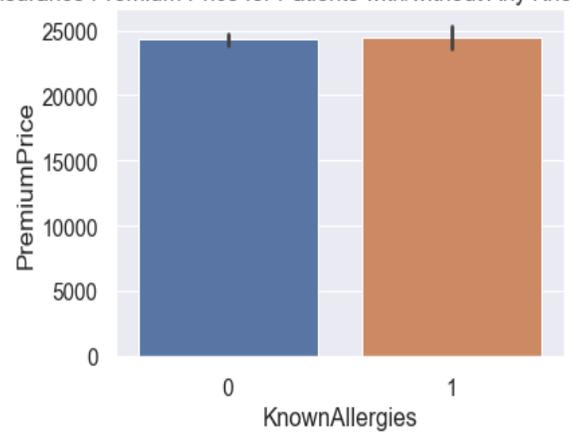
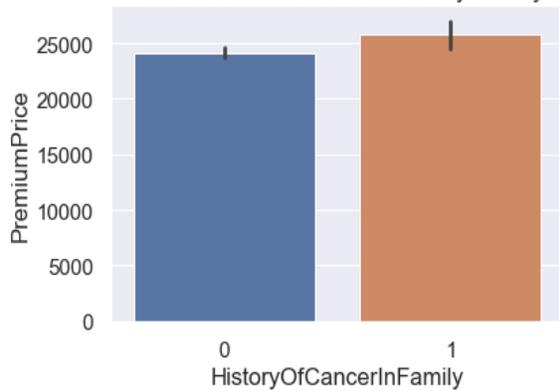

*Figure 4a: Premium Price vs. Various Independent Variables*

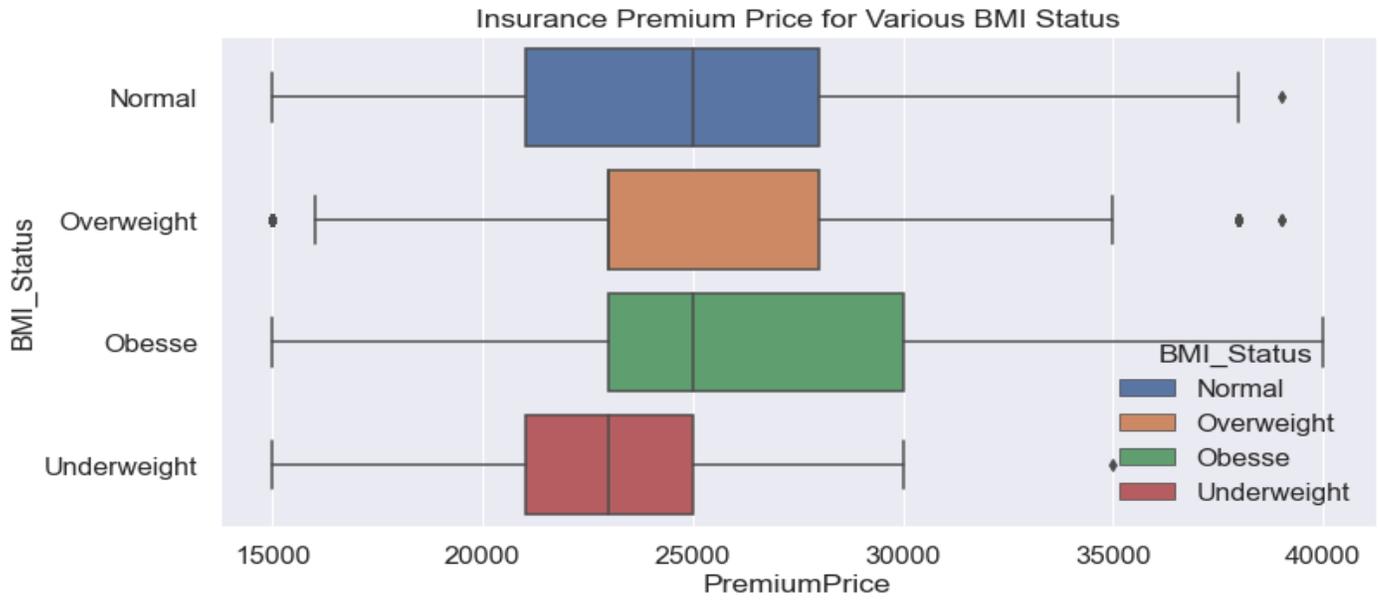

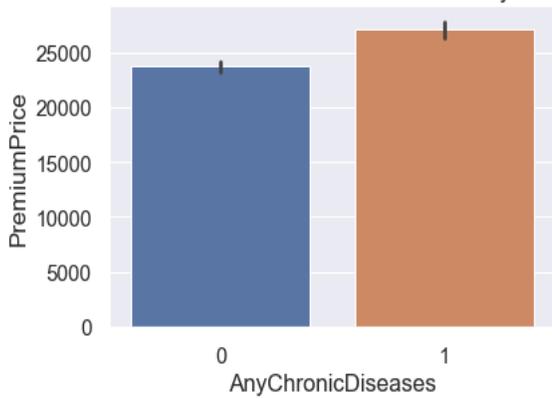
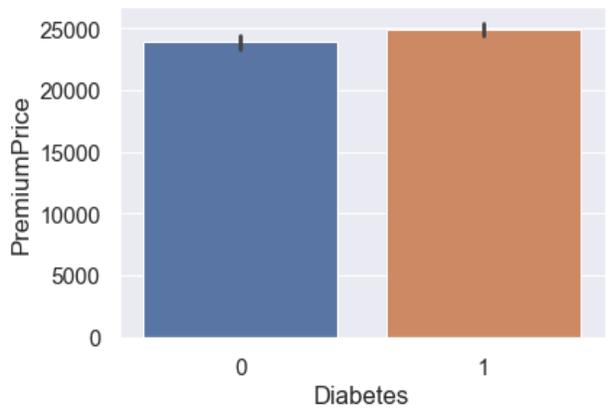

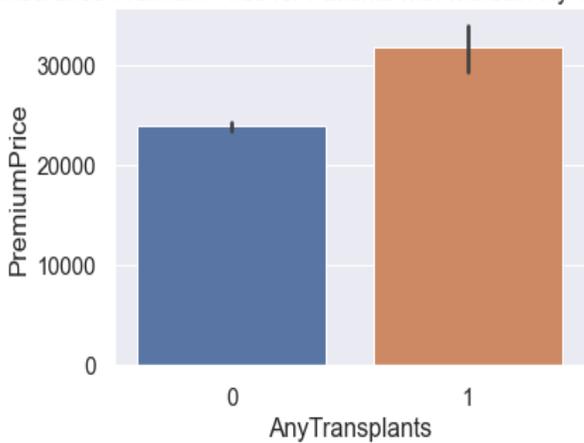
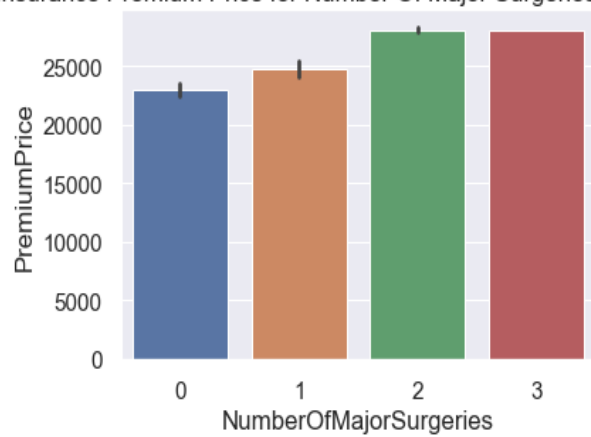

*Figure 4b: Premium Price vs. Various Independent Variables*

## 3.4 Machine Learning Models

After preprocessing, all the features were selected for the experiment. Thereafter, the data was split into two parts: the training and test datasets. Approximately 75% of the total data is used to train, and 25% for testing. As a result of the training dataset, models were created that predict medical insurance costs, whereas the test dataset was used to evaluate the regression models.

Three ensemble regression models were created through the scikit-learn's XGBoostRegressor, GradientBoostingRegressor, and RandomForestRegressor. Ensemble ML methods are known to offer the possibility of averaging many trees and overcoming the instability associated with a single model, which according to Duncan *et al.* (2016), makes ensemble models a promising choice for healthcare predictive modeling. In the following section, a brief description of each model deployed is provided.

### 3.4.1 eXtreme Gradient Boosting (XGBoost) Model

The XGBoost model was developed to become an efficient and scalable implementation of the gradient boosting framework, which was initially designed by Friedman *et al.,* (2000). XGBoost aims to improve model performance and computational speed via the boosting approach (Mitchell & Frank, 2017; Ogunleye & Wang, 2019).

*Table 4: XGBoost Model Description*

| SYMBOL | MEANING |
|---|---|
| $n$ | Total number of samples |
| $m$ | Total number of features |
| $x_i$ | Features$_i$ information of the i-th sample, xi $\in R^m$ |
| $y_i$ | The actual label (or value) of the i-th sample |
| $\hat{y}_i$ | The predicted label (or value) of the i-th sample |
| $\hat{y}_i^{(t)}$ | The predicted value up to the t-th tree |
| $l(y_i, \hat{y}_i)$ | The loss function of the i-th sample |
| $L(y, \hat{y})$ | The loss function of the total sample |
| $\Omega(f_k)$ | The regular term of the objective function to prevent overfitting, where f$_k$ represents the *k*-th decision tree. |

Given a data set containing $n$ samples and $m$ features, $D = \{(x_i, y_i) \mid x_i \in R^m, y_i \in R\}$ and $x_i = \{x_{i1}, x_{i2}, \cdots, x_{im} \mid i = 1,2,\cdots, n\}$. XGBoost will build *t* trees such that the predicted value $\hat{y}_i^{(t)}$ up to the t-th tree satisfies equation (1).

$$\hat{y}_i^{(0)} = 0$$
$$\hat{y}_i^{(1)} = f_1(x_i) = \hat{y}_i^{(0)} + f_1(x_i)$$
$$\hat{y}_i^{(2)} = f_1(x_i) + f_2(x_i) = \hat{y}_i^{(1)} + f_2(x_i) \qquad \text{Equation 1}$$
...
$$\hat{y}_i^{(t)} = \sum_{k=1}^{t} f_k(x_i) = \hat{y}_i^{(t-1)} + f_t(x_i)$$

For each iteration of the algorithm, a weak classifier $f_k(x_i)$ is generated, and the predicted value $\hat{y}_i^{(t)}$ of each iteration is the sum of the predicted value of the previous iteration $\hat{y}_i^{(t-1)}$ and the decision tree result of this round $f_t(x_i)$ (Li *et al.*, 2020).

According to Sheridan *et al.,* (2016), predictions by the XGBoost model are on average better than those of the RF model and comparable with those of deep neural nets.

Furthermore, in contrast to simple gradient boosting algorithms, XGBoost does not add weak learners one after another instead, the algorithm employs a multi-threaded approach, ensuring that the machine's CPU core is properly utilized and as such, improves overall performance and speed.

### 3.4.1.1 Properties of the XGBoost method
1. **Speed**: In general, this method has been observed to be 10 times faster than gradient boosting for both Windows and Linux Operating systems (Chen, *et al.*, 2015).
2. **Input Type**: the XGBoost model is renowned for accepting a variety of input data, e.g. Dense and Sparse Matrix, Local Data Files, etc.
3. **Highly Customisable**: XGBoost is known for its flexibility thus, its objective function and evaluation function are customisable.
4. **Performance**: XGBoost is a high-performance algorithm and has been known to perform well in a variety of datasets.

### 3.4.2 Gradient Boosting Machine (GBM)
GBMs are a family of powerful machine-learning techniques that have shown considerable success in a wide range of practical applications (Natekin & Knoll, 2013). GBM is known for its high predictive accuracy and effective handling of complex data relationships (Luo *et al*., 2019). It is widely used in various applications, including Kaggle competitions and real-world problems where accurate predictions are essential.

The mathematical framework of GBM and XGBoost is similar, as XGBoost is a specific implementation of the GBM algorithm with some enhancements, so equation 1 applies.

### 3.4.2.1 Properties of the GBM Model
1. Parallelization: GBM can be parallelized to use multiple CPU cores, making training on large datasets efficient (Atweh *et al*., 2018).
2. Customizable Loss Functions: Depending on the implementation, GBM allows you to define custom loss functions, making it adaptable to various problem types and objectives.
3. Consistency: GBM produces consistent and reliable results, making it a popular choice for many data science tasks (Konstantinov *et al*., 2021).

### 3.4.3 Random Forest (RF) Model
The RF algorithm is a tree-based ensemble algorithm that uses decision trees as base learners (Breiman, 2001). In this model, each tree is dependent on a set of random parameters. The RF trees are constructed from bootstrap samples of the primary dataset.

In comparison to single decision tree methods, the RF model is superior due to their ability to reduce over-fitting by averaging the outcome. There are several steps to the RF technique including:

- First, the bootstrap samples are randomly sampled from the original data set (Bagherzadeh & Shafighfard, 2022).
- Afterward, a decision tree is built for each of the samples and an output is produced.
- Subsequently, voting is used to decide on the most popular predictive model which serves as the final forecast (Breiman, 2001; Cutler, Cutler & Stevens, 2012; Heddam, 2021).

RF prediction is the unweighted average over the collection according to Segal (2004) as given below:

$$h(\boldsymbol{x}) = (1/K) \sum_{k=1}^{K} h(\boldsymbol{x}; \theta_k) \qquad \text{Equation 2}$$

As k → ∞ the Law of Large Numbers ensures

$$E_{X,Y}(Y - h(\boldsymbol{X}))^2 \to E_{X,Y}(Y - E_\theta h(\boldsymbol{X}; \theta))^2 \qquad \text{Equation 3}$$

Where $h(\mathbf{x}; \theta_k)$, $k = 1,\ldots, K$ is a collection of tree predictors, **x** represents the observed input vector of length *p* with associated random vector **X** and the θ$_k$ are independent and identically distributed random vectors.

### 3.4.3.1 Properties of the RF Model

Although the RF model is relatively complex to interpret per Ziegler & König (2014), researchers have extensively studied the theoretical properties of RFs see Biau & Scornet (Biau & Scornet, 2016). Some properties of the RF model include:

1. RF models are known for their consistency (Tyralis, Papacharalampous & Langousis, 2019; Scornet, Biau & Vert, 2015).
2. RF models have been known to reduce variance while not increasing the bias of the predictions (Genuer, 2012).
3. RF models can easily reach a minimax rate of convergence (Ziegler & König, 2014; Genuer, 2012).
4. RF models are easily adaptable and can handle missing values (Tyralis, Papacharalampous & Langousis, 2019).

*Table 5: Advantages and Disadvantages of the ML Models*

| Models | Advantage | Disadvantage |
|---|---|---|
| RF Model | RF ensures the availability of reliable estimates for feature importance (Zhao *et al.,* 2022). | RF models have longer computation time and consume more computational resources (Biau & Scornet, 2016). |
| | RF model performs well even without hyper-parameter tuning (Gomes *et al.,* 2017). | Prone to overfit with noisy data (Hoarau *et al.,* 2023). |
| | The presence of missing values does not hinder RF (Tyralis *et al*., 2019). | RF is relatively hard to interpret (Marchese Robinson *et al.,* 2017). |

| XGBoost Model | XGBoost performs well with little or no feature engineering and can handle missing data (Kang *et al.*, 2020). | If not properly tuned, XGBoost is more likely to overfit (Priscilla & Prabha, 2020). |
|---|---|---|
| | XGBoost is renowned for its computational speed, model performance, and is well-known to handle large-sized datasets (Chen, *et al.*, 2015). | It is harder to tune as there are too many hyper-parameters (Zivkovic *et al.*, 2022). |
| GBM Model | Improved convergence speed without a significant decrease in accuracy (Feng *et al.,* 2018) | Achieving a balance between performance and generality has posed a challenge for GBMs (Luo et al., 2022). |
| | Gradient boosting of regression trees produces competitive, highly robust, interpretable procedures for both regression and classification, especially appropriate for mining less than clean data (Friedman, 2001). | Like in XGBoost, GBM has many hyperparameters that need proper tuning (Anghel *et al.,* 2018; Kiatkarun & Phunchongharn, 2020). |

### 3.5 Model evaluation approach

Model evaluation is an important aspect of building ML projects as it helps understand the model performance and makes it easier to explain/present the result of the models. However, it can be difficult to predict the exact value of a regression model; thus, the goal becomes to show how close the predicted values are to the actual values. In this research, we evaluated the models using 4 performance evaluation metrics: $R^2$, MAE, RMSE and MAPE.

$R^2$ is the square of the Correlation Coefficient (R), it is used for evaluating the fitting of a model, and more specifically for examining the proportion of the predicted price that can be explained by the features. Mean Absolute Error (MAE) is calculated to accurately reflect prediction errors. Root Mean Squared Error (RMSE) measures the standard deviation of residuals, it tells how well a regression model can predict the value of a response variable. Both MAE and RMSE give the error in terms of the value being predicted, in our case for example, MAE and RMSE would give the error in terms of premium prices. Finally, the Mean Absolute Percentage Error (MAPE) computes errors in terms of percentage; more specifically, it is the average percentage difference between predictions and their intended targets in the dataset. MAPE can also be seen as the MAE returned as a percentage.

According to Allwright (2021), a reading below 10% for MAPE reflects high-quality predictive modeling. Furthermore, for an ideal model, an $R^2$ score near 100% is the best possible outcome and suggests the more accurate and better performing the model is.

We can illustrate these metrics as follows:

Let $n$ denote the number of samples, $y_i^p$ denote the *i*th prediction of the sample, $y_i$ is the corresponding actual value and reflects the mean value of the sample.

$$R^2 = 1 - \frac{\sum_{i=1}^{n}(y_i - y_i)^2}{\sum_{i=1}^{n}(y_i - \bar{y})^2}, \text{ where } \bar{y} = \frac{1}{n}\sum_{i=1}^{n} y_i \qquad \qquad \text{\textit{Equation 4}}$$

$$MAE = \frac{1}{n}\sum_{i=1}^{n} |(y_i - y_i^p)| \qquad \text{Equation 5}$$

$$RMSE = \sqrt{\frac{1}{n}\sum_{i=1}^{n} (y_i - y_i^p)^2} \qquad \text{Equation 6}$$

$$MAPE = \frac{1}{n}\sum_{i=1}^{n} \left|\left(\frac{y_i - y_i^p}{y_i}\right)\right| * 100 \qquad \text{Equation 7}$$

## 3.6 Explainable AI (XAi)

Generally, ML models have shown to be reliable and efficient, but their superiority comes at the cost of algorithm complexity making them black-box models (Guidotti, 2018). It can be challenging to quantify the contribution of each feature to the predicted outcome of an ML model (Friedman, 2001). Accuracy versus interpretability has remained a major issue at the intersection of ML and the explainability of models (Gunning, 2019). To overcome this challenge, the Defense Advanced Research Projects Agency (DARPA) developed the XAI program aimed at making ML models understandable, trustworthy, and transparent to end users (Gunning, 2021; Adadi & Berrada, 2018).

The ML explainability concept sits at the intersection of several areas of active research in AI, with a focus on the following areas according to Hagras (2018):

- **Transparency**: the need to understand the actual mechanism of ML models (Weller, 2017).
- **Causality**: the need for explanations from ML models like their underlying phenomena.
- **Bias**: the need to ensure ML models are unbiased from shortcomings of the training data or other factors.
- **Fairness**: the need to ensure fairness in AI systems
- **Safety:** ensuring the reliability and integrity of ML models.

### 3.6.1 SHapley Additive exPlanations (SHAP)

SHAP was proposed by Lundberg & Lee (2017) as a method for providing insight into the mechanisms of a specific instance by measuring each feature's contribution. SHAP is now widely used to interpret natural and social phenomena (Stojić, *et al.,* 2019; Janizek, Celik & Lee, 2018). The idea behind SHAP is grounded in game theory and its axioms - such as assigning zero attributions to features that did not contribute to the model prediction.

Consequently, SHAP aims to explain a model *f* at some individual point *x\** with a value function $e_S$ where *S* is a subset of $S \subseteq \{1,..., p\}$. Typically, this function is defined as the expected value for a conditional distribution as applied to every variable *S* and given as:

$$e_S = E[f(x) \mid x_S = x_S^*] \qquad \text{Equation 8}$$

As described by Holzinger *et al.,* (2022), the contribution of a variable *j* is denoted by $\phi_j$ and calculated as the weighted average over all possible *S* as shown below:

$$\phi_j(val) = \sum_{S \subseteq \{x_1,...,x_p\} \setminus \{x_j\}} \frac{|S|!(p-|S|-1)!}{p!} \left(val\left(S \cup \{x_j\}\right) - val(S)\right) \qquad \text{Equation 9}$$

where *p* denotes the number of features, *S* is a subset of the features, *x* represents the vector of feature values of an instance in the model that is being explained and *val(S)* denotes the prediction for feature values in the set *S*.

SHAP considers the contribution of each determinant by computing all the different permutations of the explanatory variables and evaluating the contribution of each one. It then takes an average of those contributions to get the final contribution of the feature (Teoh, *et al.,* 2022).

Additionally, SHAPley values can be investigated in more detail to obtain global interpretations of the model, such as feature dependence plots, feature importance plots, and interaction analysis (Holzinger, 2022). To calculate the absolute SHAPley values per characteristic the formula is given below:

$$I_j = \sum_{i=1}^{n} \left| \phi_j^{(i)} \right| \qquad \qquad Equation\ 10$$

Where $\phi_j^{(i)}$ represents the SHAP value of the *j-th* feature for instance *i*.

A spatial variation of each characteristic's contribution and complex nonlinear relationships between each feature in the medical insurance cost dataset can be explained by the SHAP method. To explore the relationships between premium prices and other features of the dataset, we used SHAP beeswarm/summary plots.

### 3.6.2 Individual Conditional Expectation (ICE) Plots

As reported by Goldstein *et al.* (2015) ICE plots originated from Friedman's Partial Dependence Plot (PDP), which highlights an average partial relationship between a set of predictors and the predicted response (Friedman, 2001). As opposed to a PDP, an ICE plot shows the actual prediction functions $\hat{f}^{(i)}$ (i.e. the "individual" conditional expectations), rather than the averages. Visually, ICE plots are a visual disaggregation of classical PDPs. ICE plots can reveal the partial dependence of a feature for each instance separately and the average of all ICE curves is equivalent to the PDP (Zhang, 2022; Molnar, 2020).

The ICE plot illustrates the observed range of an independent variable along the x-axis and the predicted value/probability on the y-axis. Each line illustrates one individual expectation (Jordan, Paul, & Philips, 2022a).

We consider the set of observations $\{(x_{Si}, \boldsymbol{x}_{Ci})\}_{i=1}^{N}$, and examine the estimated response function $\hat{f}$. For each observed value *N* and fixed values of $\boldsymbol{x}_C$, a subsequent curve given as $\hat{f}_S^{(i)}$ is drawn against the observed values of $x_S$. Therefore, at each x-coordinate, $x_S$ is fixed, and the $\boldsymbol{x}_C$ varies across *N* observations. Each curve is used to equate the conditional relationship between $x_S$ and $\hat{f}$ at $\boldsymbol{x}_C$. Through the use of this ICE algorithm, we gain insight into various conditional dependencies estimated by the black box model.

Two extensions of ICE plots exist, one is a "centered" ICE plot, or c-ICE (Goldstein *et al.* 2015, Jordan, Paul, & Philips, 2022a), whereby some observed location $x^*$ along $x_S$ is chosen, and all lines are forced to run through that point. The c-ICE plots for plotting along a single predictor variable are given as:

$$\hat{f}_{cICE}^{(i)} = \hat{f}^{(i)} - \mathbf{1}\hat{f}(x^*, x_{Ci}) \qquad \text{Equation 11}$$

where $\hat{f}^{(i)}$ is the original ICE curve and $\hat{f}(x^*, \mathbf{x}_{Ci})$ is the prediction for location $x^*$ for observation $i$.

The second known as Derivative ICE (d-ICE) plots show the partial derivative of each ICE curve with respect to the predictor variable of interest and can be used to probe for 'hidden' interactions between this predictor and the other control variables. If there are no interactive effects, the d-ICE plot will look like a single line that shows the partial derivative (i.e., the effect is constant across all observations). However, if there are interactive effects, they will appear as heterogeneous partial derivative lines in the plot. Derivatives cannot be analytically derived but are instead numerically approximated according to (Goldstein et al. 2015); therefore, creating d-ICE plots often takes a long time.

To illustrate, consider the scenario in which $x_S$ does not interact with the other predictors in the fitted model. Thus $\hat{f}$ can be written as:

$$\hat{f}(\mathbf{x}) = \hat{f}(x_S, \mathbf{x}_C) = g(x_S) + h(\mathbf{x}_C), \text{ so that } \frac{\partial \hat{f}(x)}{\partial x_S} = g'(x_S) \qquad \text{Equation 12}$$

This implies the relationship between $x_S$ and $\hat{f}$ does not depend on $\mathbf{x}_C$. Therefore, the ICE plot for $x_S$ would display a set of $N$ curves that share a single common shape but differ by level shifts according to the values of $\mathbf{x}_C$.

According to Jordan, Paul, & Philips (2022b), ICE plots are best suited where a heterogeneous relationship between the features is expected; thus it is suitable for this study.

*Table 6: Advantages and Disadvantages of SHAP*

| XAI Method | Advantage | Disadvantage |
|---|---|---|
| SHAP Analysis | Global interpretability— SHAP helps determine whether each variable is positively or negatively related to the target variable (Lundberg & Lee, 2017). | SHAP comes with computational complexity and consumes huge computation resources (Lin & Gao, 2022). |
| | Local interpretability— all features are represented with a SHAP value (Stiglic et al., 2020). | SHAPley values cause extrapolation to low-density areas for dependent features (Lundberg & Lee, 2017). |
| | SHAP calculates the contribution of each feature to the prediction (Teoh, *et al.*, 2022). | Regardless of how small the change may be, every feature that changes the prediction is attributed a SHAPley value other than zero (Janizek, Celik & Lee, 2018). |
| ICE Plots | A fitted model's ICE plot can reveal heterogeneous relationships between predictors and predicted values by visualizing the map between predictors and predicted values (Casalicchio et al, 2019). | According to the joint feature distribution, some points on the ICE curves might be invalid data points if the feature of interest is correlated with the other features (Molnar *et al*. 2022). |

| | The process of creating an ICE plot is extremely straightforward (Goldstein et al. 2015). | Generating ICE plots can be time-consuming, especially for large datasets or complex models (Molnar, 2020). |

# 4 Results and Discussion

## 4.1 Hyperparameter Settings and GridSearchCV

Hyperparameters play a crucial role in model development and performance. Well-tuned models perform better (Alhakeem *et al.,* 2022). Grid Search Cross-Validation (GridSearchCV) was used to choose the best parameters for each ML approach. GridSearchCV is a well-known method for determining all hyperparameter combinations and helps in obtaining a more accurate generalization performance estimate by ensuring optimal values for a model's hyperparameters are used (Ahmad *et al*., 2022). A collection of values is initially determined for each hyperparameter at each cycle or iteration. Finally, the most successful hyperparameter combination is selected and implemented in the learning process.

The learning rate and the number of estimators are some of the most important parameters in GridSearchCV (Sumathi, 2020). Max_features denotes the maximum number of variables used in independent trees, while n_estimators are the total number of trees constructed. Table 7 provides an overview of the model training result, computational time and memory utilization for the models, and parameter tuning in our experiment.

*Table 7. Model training score and GridSearchCV*

| Model | Train Score ($R^2$) | Elapsed Time (Secs.) | Memory Used (MB) | Tuning Parameters | $R^2$ Score (5-fold Cross Validation) | Best Parameters |
|---|---|---|---|---|---|---|
| **RF** | 95.809 | 55.060 | 0.922 | n_estimators: [60, 220, 40] max_depth: [7] min_samples_split: [3] max_features: ['auto'] | 73.428 | n_estimators: 220 max_depth: 7 min_samples_split: 3 max_features: 'auto' |
| **XGBoost** | 88.222 | 5272.079 | 6.660 | max_depth: range (2, 10, 1) n_estimators: range (60, 220, 40) learning_rate: [0.1, 0.01, 0.05] subsample: [0.6,0.7,0.75,0.8,0.85,0.9] | 74.475 | {'gamma': 0, 'learning_rate': 0.1, 'max_depth': 5, 'n_estimators': 50, 'subsample': 0.9} |
| **GBM** | 79.842 | 58.408 | 0.000 | n_estimators:[10,15,19,20,21,50,100] learning_rate:[0.1,0.19,0.2,0.21,0.8,1] | 72.999 | {'learning_rate': 0.19, 'n_estimators': 19} |

Overall, the XGBoost model expanded more computational resources (time and memory) than the GBM and RF models as shown in Table 7.

Furthermore, the learning curve graph was used to visually represent how the models' performance improves or stabilizes as it learns from more training examples. It shows how the models' accuracy changes as the size of the training dataset increases. This curve helps us understand how well the model generalizes from the data and whether it might suffer from underfitting or overfitting.

Learning curves often include two lines: one representing the training performance and the other representing the validation (or cross-validation) performance. The validation curve helps us assess how well the model generalizes to new, unseen data. An ideal learning curve shows both the training and validation curves improving and stabilizing at high levels. This indicates that the model is learning from the data effectively and generalizing well to new instances.

The x-axis of the learning curve represents the number of training instances (data points) used to train the model, while the y-axis represents a performance metric, in our case, the accuracy ($R^2$) as given in *Equation 4* to represent how well the model is doing.

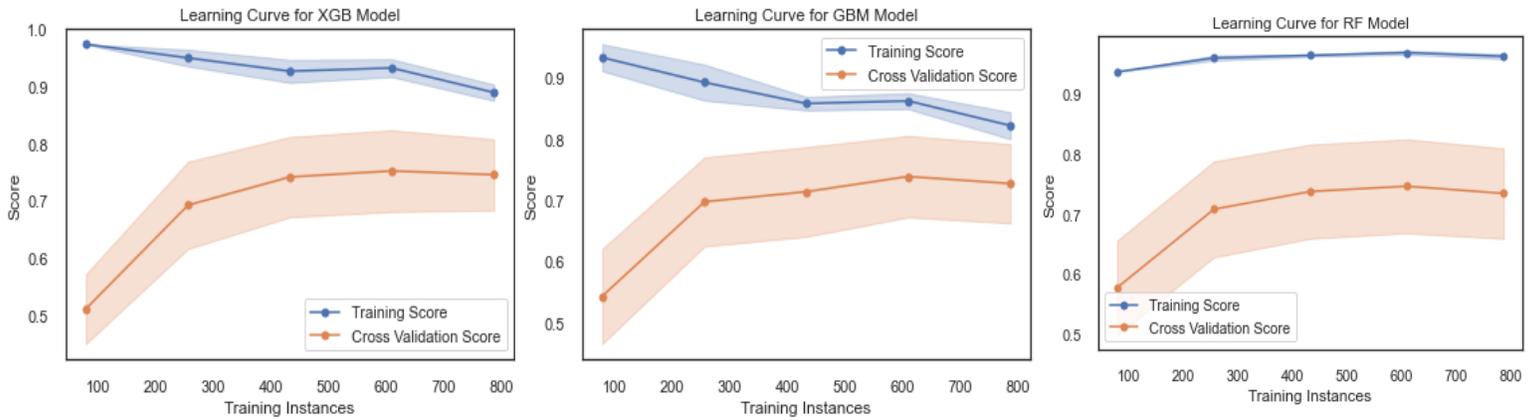

*Figure 5: Learning curve for XGBoost, GBM, and RF Models*

As shown in Figure 5, for all the models from the 600 training instance, it can be observed that the learning curve starts to plateau, indicating that adding more training data does not lead to significant improvements in performance. This indicates that the model is converging and has learned the majority of the patterns present in the data.

Furthermore, to improve the results, we apply the best parameters from the GridSearchCV, as shown in Table 7 to the test data and confirm the outcome, as shown in Table 8 below.

*Table 8: Percentage increment due to applying hyperparameter optimization*

| ML Method | $R^2$ (%) Train data | $R^2$ (%) on Cross Validation data | $R^2$ (%) Test data | % improvement after applying hyperparameters |
|---|---|---|---|---|
| **XGBoost** | 88.222 | 74.475 | 86.470 | 14.906 |
| **RF** | 95.809 | 73.428 | 84.046 | 13.485 |
| **GBM** | 79.842 | 72.999 | 84.720 | 14.863 |

The results in Table 8 show that all the models had a considerable improvement from the outcome of the cross-validation. It can be observed that the XGBoost and GBM models had a higher improvement than the RF model after hyperparameters were applied; this could be attributed to the gradient-boosting framework, which is particularly effective for optimization tasks via its regularization terms, and sophisticated tree-building techniques. These elements work together to iteratively improve the models' predictions, reduce overfitting, and achieve high performance.

Also, while the RF algorithm itself is not designed for hyperparameter optimization; however, having identified the important parameters that needed tuning, as seen in Table 7, the model also achieved better performance and generalization on the test data.

## 4.2 Summary of Model result on test data

*Table 9: Model performance result on test data*

| ML Method | $R^2$ (%) | MAE | RMSE | MAPE (%) |
|---|---|---|---|---|
| **XGBoost** | 86.470 | 1442.904 | 2231.524 | 5.906 |
| **GBM** | 84.720 | 1725.859 | 2371.412 | 7.229 |
| **RF** | 84.046 | 1379.960 | 2423.161 | 5.831 |

The prediction result shown in Table 9 indicates that the models produced impressive outcomes. However, the XGBoost model achieved the best outcome with an $R^2$ score of 86.470% and an RMSE of 2231.524. Comparatively, RMSE penalizes large gaps more harshly than MAE; here, our result shows that the RF model obtained lesser MAE and MAPE than the GBM and XGBoost models. The GBM model made more large-scale prediction errors than the XGBoost and RF models as shown by its high MAE and MAPE scores. Also, the high $R^2$ score of the XGBoost model can be explained by its ability to explain the variation in the data better than the GBM and RF models. Considering that the mean PremiumPrice from the dataset is 24336.71, as seen in Table 3, we conclude that all the models achieved impressive prediction results.

Furthermore, we plotted the residual of the models as shown in Figures 7, 8, and 9. The residuals plot visualizer lets you identify which regions of the target variable are prone to errors by plotting the difference between the residuals and the target variable (*y*) on a vertical and horizontal axis.

Here the residual plot reveals that all the models are well-balanced, with a symmetric distribution around the middle of the plot. The quantile-quantile (*Q-Q*) plot further reinforces this observation, demonstrating normal data skewness and model fit for all models.

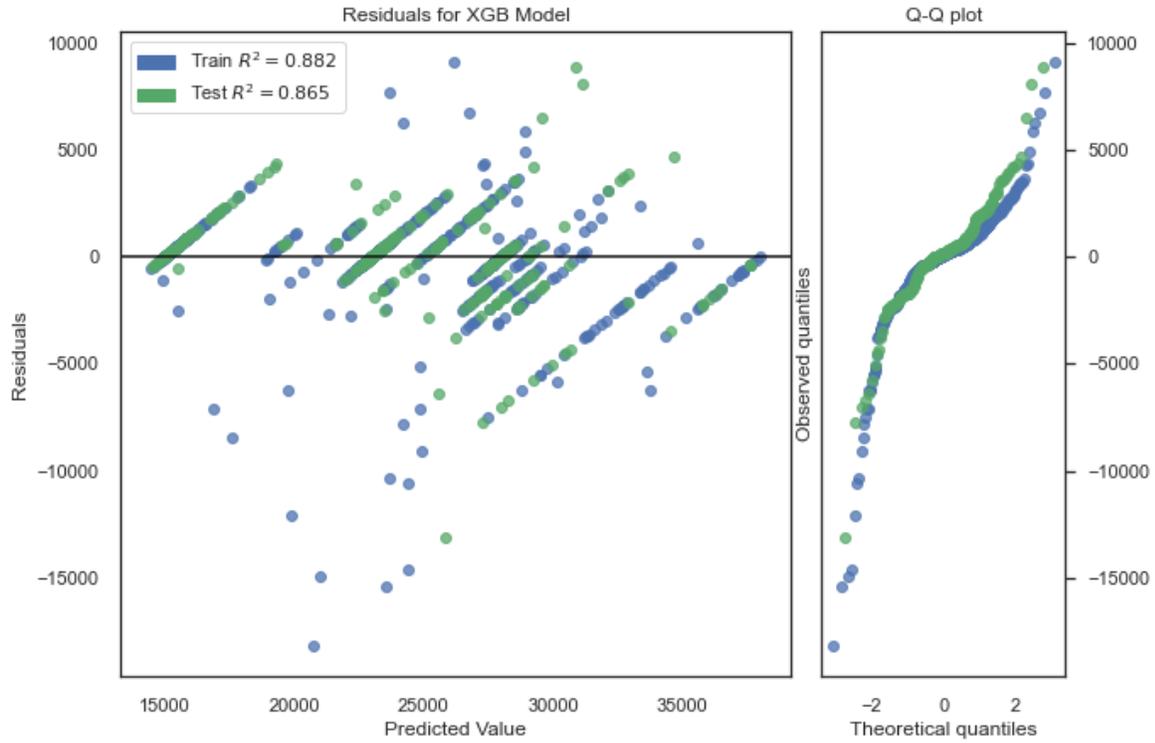

*Figure 7: XGBoost model residual plot*

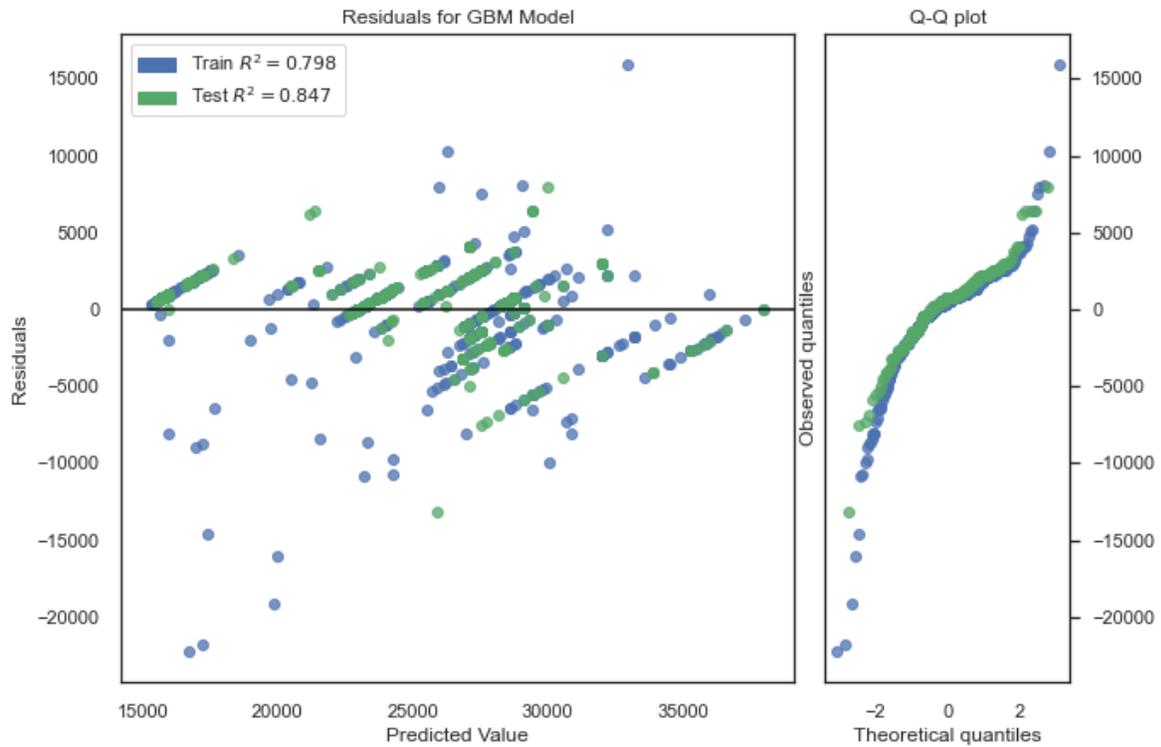

*Figure 8: GBM model residual plot*

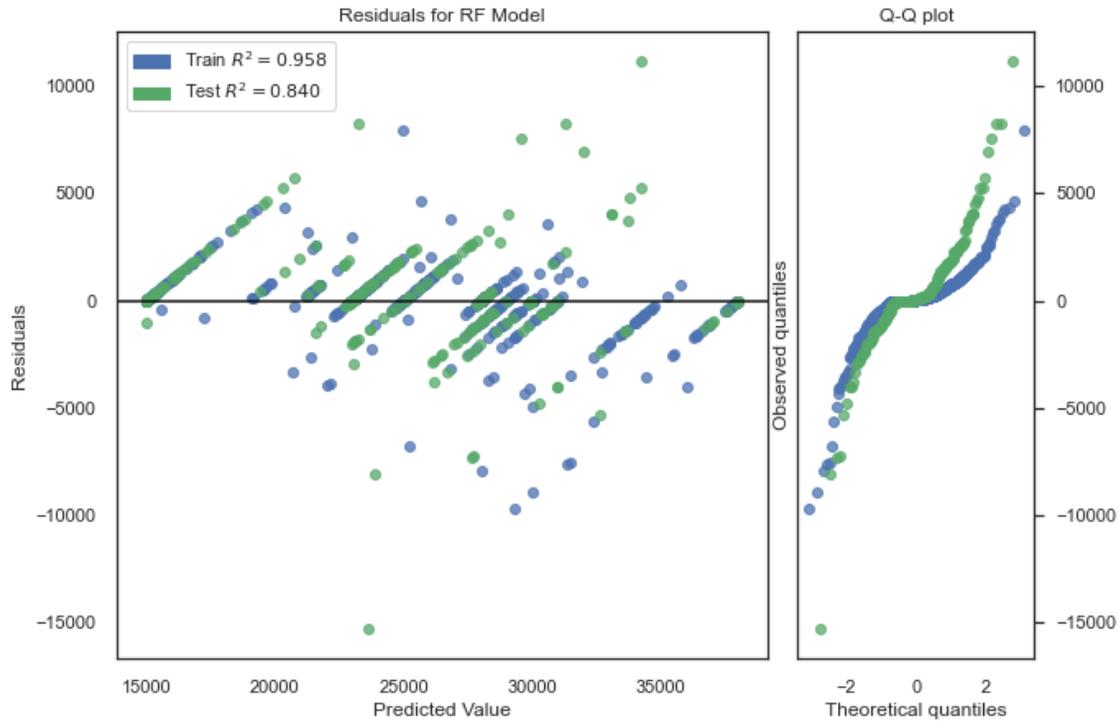

*Figure 9: RF model residual plot*

Furthermore, the actual targets from the dataset were plotted against the predicted values generated by the model using Yellowbrick's Prediction Error Visualizer. The plot shows how much variance there is in the model. The authors compared this with the 45-degree line, which exactly matches the model as shown in Figure 10 below.

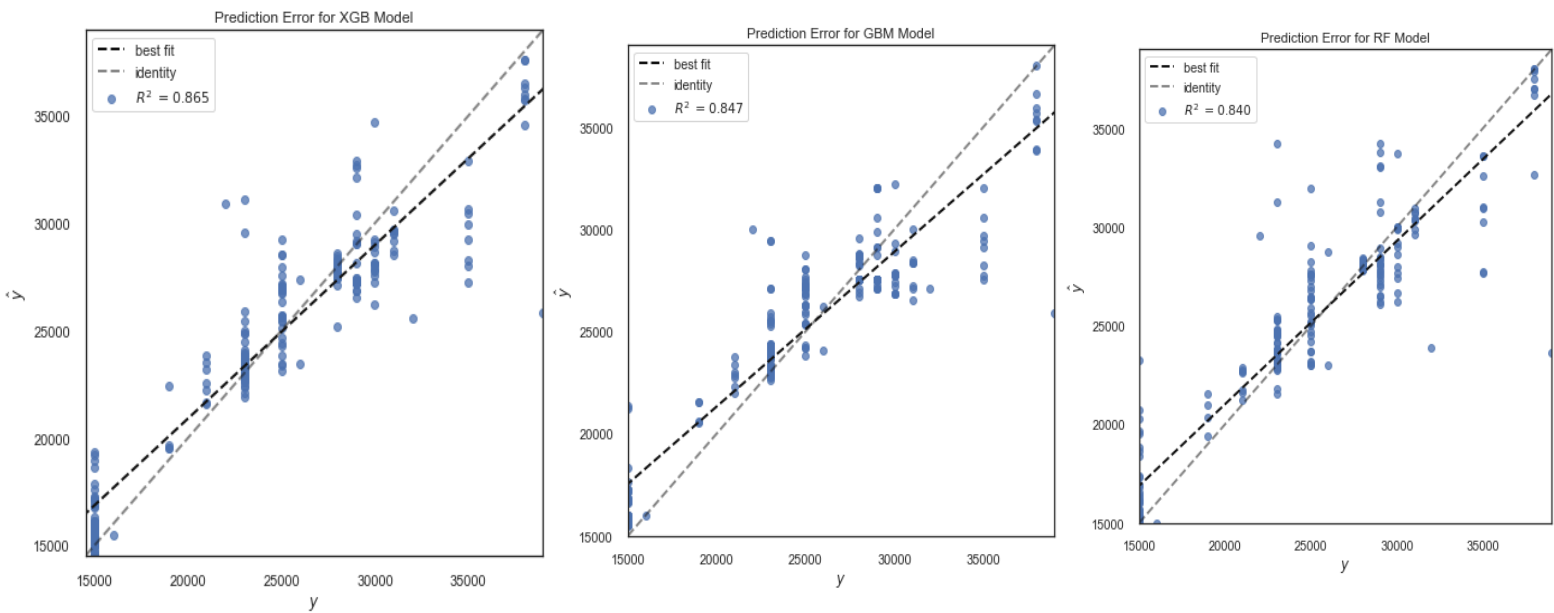

*Figure 10: Prediction error plot for XGBoost, GBM, and RF Models*

## 4.3 Summary of Determinant Analysis using SHAP

The SHAP beeswarm or summary plot is a great tool for visualizing the outcomes of a model. It provides information on how different features in the dataset affect model output, as represented by SHAPley values coupled with feature importance and effects. The *y*-axis of the plot reflects the feature, and the *x*-axis reflects a SHAPley value which in our case ranges from -10,000 to 10,000. The SHAPley value range is bounded by the output magnitude range of the model (i.e. PremiumPrice). Thus, it explains the impact of each feature on the PremiumPrice.

Additionally, color coding aids in showing low to high values for each feature when compared to other observations; all the little dots on the plot represent a single observation. To better illustrate overlapping points, they may be shifted along the *y*-axis. Finally, the features are listed according to their mean SHAP and importance.

In Figure 11, one can start to understand the nature of these relationships. According to the beeswarm plot for the XGBoost model, it becomes clear that as the feature values for "Age", "BMI", "AnyTransplant", "AnyChronicDiseases", "HistoryOfCancerInFamily", and "BloodPressureProblems" increase, the SHAP value also increases, thus indicating that the higher the values of these features, the higher the premium price will be. One could also see that for the XGBoost model, a high value for the feature "NumberOfMajorSurgeries" seems to negatively influence the premium price, while a smaller value for the feature increases the premium price. Furthermore, the "KnownAllergies" and "Diabetes" features seemed to have a negligible influence on the premium price.

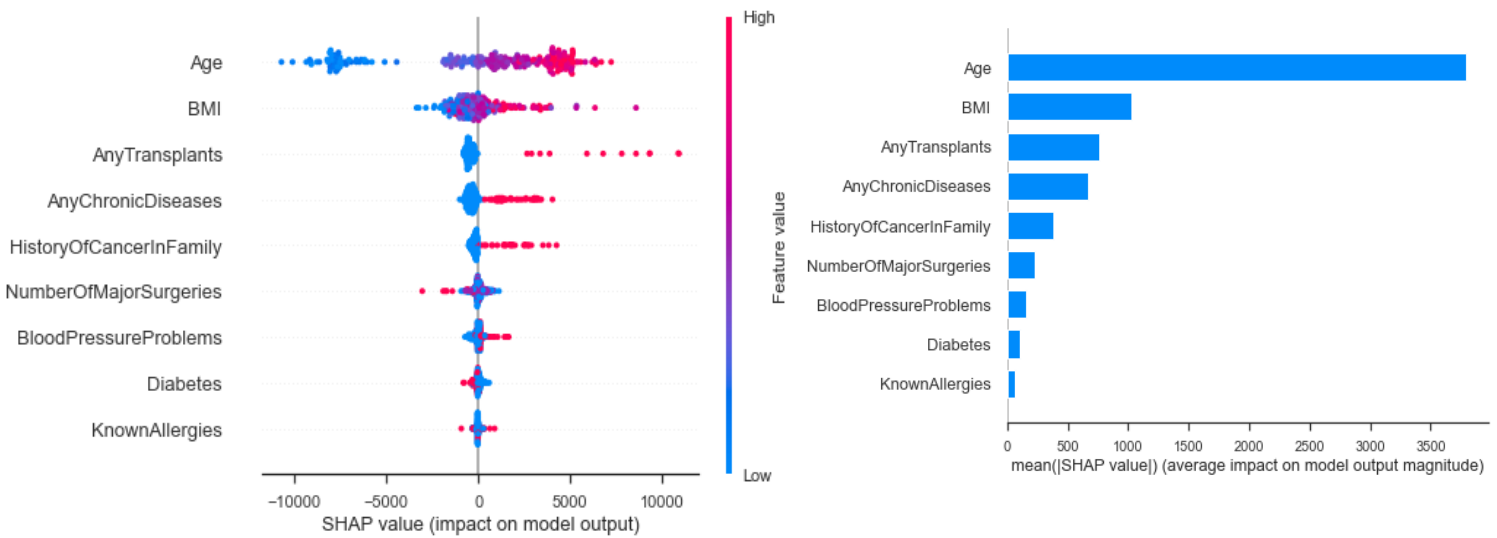

*Figure 11: The SHAP summary plot and relative importance of each feature in the XGBoost model: (a) SHAP summary plot; (b) relative importance.*

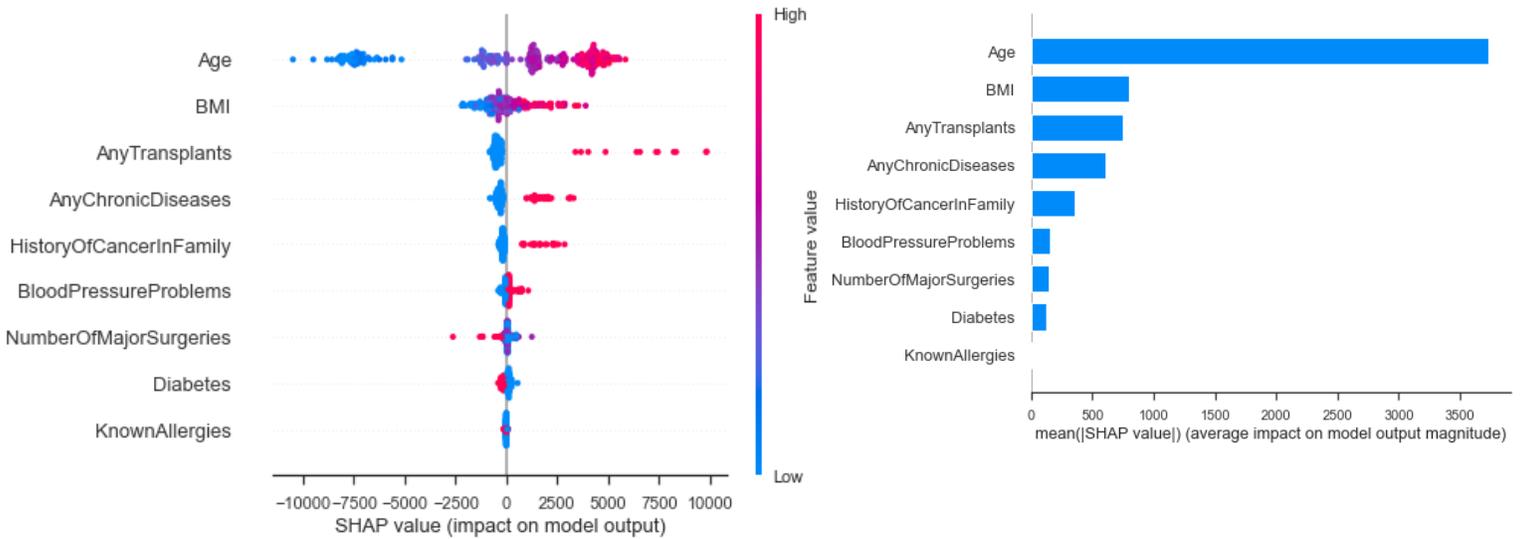

*Figure 12: The SHAP summary plot and relative importance of each feature in the GBM model: (a) SHAP summary plot; (b) relative importance.*

For the GBM model, as shown in Figure 12 above, "Age", "BMI", "AnyTransplant", "AnyChronicDiseases", and "HistoryOfCancerInFamily" are the most significant contributors to high premium prices. Thus, the higher the features, the higher the premium price, and like the XGBoost model, a high number of "NumberOfMajorSurgeries" had a negative effect on the price, while a smaller value for the feature increased the premium price. Also, "KnownAllergies" and "Diabetes" had a negligible effect on the premium price. However, unlike the XGBoost, in the GBM model, the "BloodPressureProblems" feature seemed to have lesser influence on the premium price.

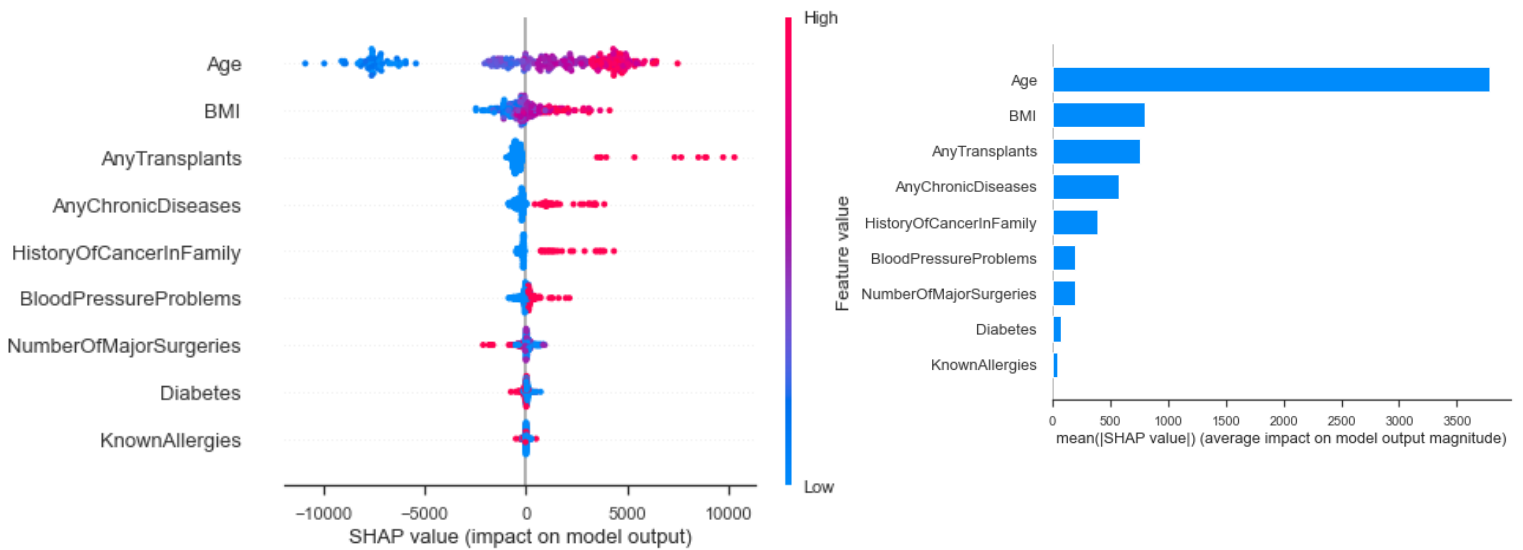

*Figure 13: The SHAP summary plot and relative importance of each feature in the RF model: (a) SHAP summary plot; (b) relative importance.*

For the RF model, the SHAP plot as shown in Figure 13, also indicates that as the feature values for "Age", "BMI", "AnyTransplant", "AnyChronicDiseases", and "HistoryOfCancerInFamily" increase, the SHAP value also increases, thus indicating that the higher the values of these features, the higher the premium price will be. The SHAP for the RF model also shows that the "BloodPressureProblems" feature had a higher influence on the premium price than seen in the XGBoost and GBM models. Again, a high "NumberOfMajorSurgeries" negatively influenced the premium price, as observed with the other models; likewise, the negligible influence of "KnownAllergies" and "Diabetes" on the premium price.

Finally, the SHAP plot for each model varies because different features made varying contributions to the outcome of each ML model deployed.

## 4.4 Summary of Determinant Analysis using ICE Plots

We employed the Centered ICE (C-ICE) plots to facilitate the comparison of individual instance curves. This approach proves valuable because it enables not only the assessment of absolute changes in predicted values but also shows the differences in predictions relative to a fixed point within the feature range. Furthermore, we added the PDP line to the plot thereby combining ICE Plots and PDPs. This helps emphasize how some of the observations deviate from the average trend.

For the XGBoost model as shown in Figure 14, it can be seen that the "Age and BMI" features had the highest interaction and were the most influential features for the price determination. Other features like "AnyTransplant and AnyChronicDiseases" showed major price increases as the feature value increased, "HistoryOfCancerInFamily and BloodPressureProblems" showed minimal increase in price as the feature values increased. For the "NumberOfMajorSurgeries" feature, it can be seen that an increase in feature value showed a decrease in price which also aligns with the observations from the SHAP. Finally, just as with the SHAP, the "KnownAllergies and Diabetes" features had negligible impact on the price determination.

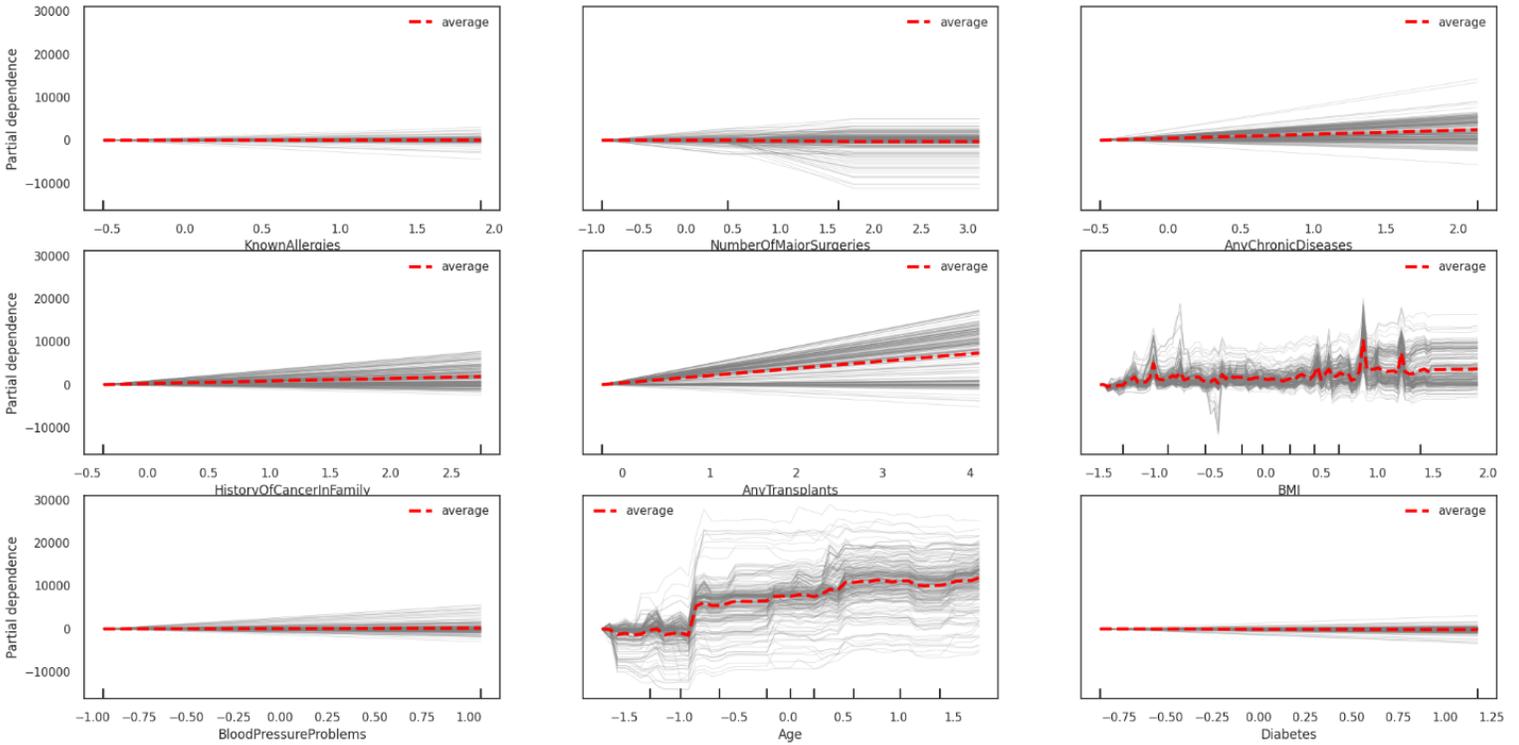

*Figure 14: C-ICE plot for XGBoost Model*

For the GBM model as shown in Figure 15, again it can be seen that the "Age and BMI" feature had the highest interaction and were the most influential features for the price determination. Other features like "AnyTransplant and AnyChronicDiseases" showed major price increases as the feature value increased, the "HistoryOfCancerInFamily and BloodPressureProblems" features also showed minimal increase in price as the feature values increased while the "NumberOfMajorSurgeries" feature showed a decrease in price when the feature values increased which also aligns with the observations from the SHAP and matches the ICE plot for XGBoost. Finally, the "KnownAllergies and Diabetes" features showed an even lesser impact on the price determination for the GBM model than the XGBoost model.

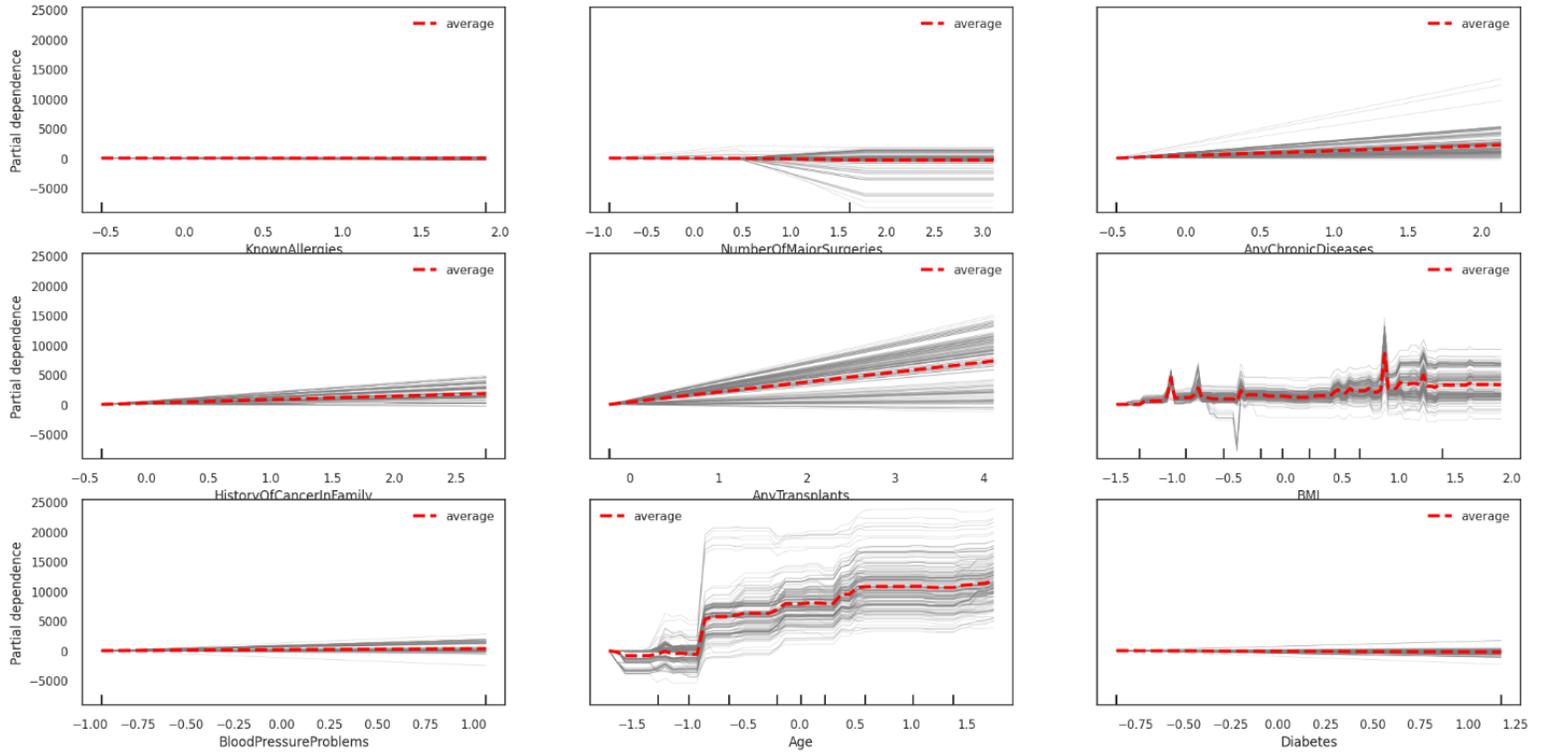

*Figure 15: C-ICE plot for GBM Model*

Figure 16 shows the C-ICE plot for the RF model, here the "Age and BMI" feature showed consistent results with the XGBoost and GBM models in producing the highest interaction and were also the most influential features for the price determination. "AnyTransplant and AnyChronicDiseases" features also showed major price increases as the feature value increased, but unlike the other two models (XGBoost and GBM), the "HistoryOfCancerInFamily and BloodPressureProblems" features showed a higher impact on the price as the feature values increased. Similar to the ICE plots for the XGBoost and GBM models, the "NumberOfMajorSurgeries" feature for the RF model also showed a decrease in price when the feature value increased. Finally, the "KnownAllergies" feature maintained a relatively negligible impact on the price while the "Diabetes" feature seemed to show a decrease in price as the feature values increased.

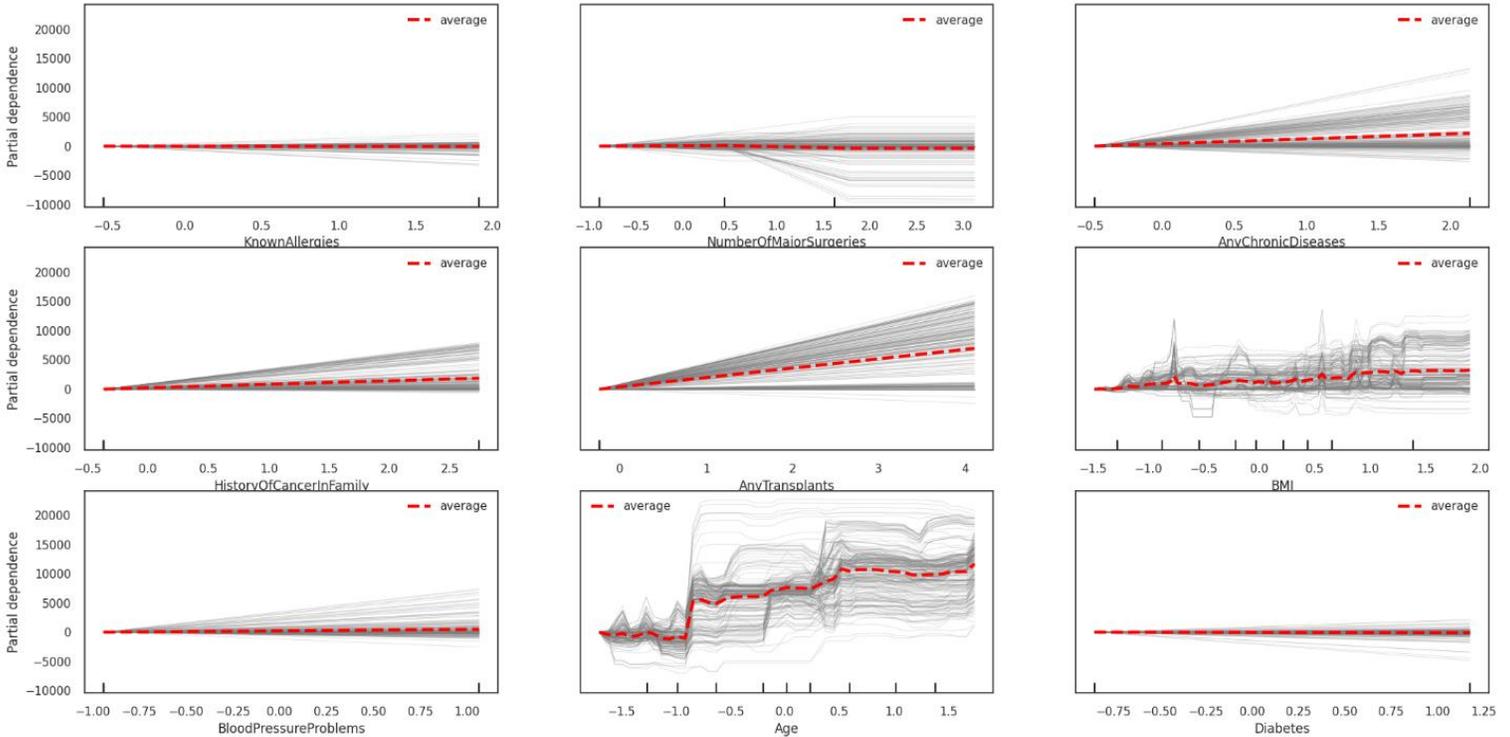

*Figure 16: C-ICE plot for RF Model*

Conclusively, the C-ICE plots for the models provide more insight into feature interactions than the SHAP.

## 5 Conclusion and future work

In this paper, three ensemble ML models, XGBoost, GBM, and RF were deployed for medical insurance cost prediction using the medical insurance cost dataset from KAGGLE's repository. The comparative result from our prediction showed that all the models achieved impressive outcomes, and whereas the XGBoost model achieved the best $R^2$ score of 86.470% and RMSE of 2231.524, it also took a more substantial computing resource to complete as shown in Table 7. The RF model achieved a better MAE and MAPE outcome of 1379.960 and 5.831% respectively and was the fastest model to build and consumed less memory than the XGBoost model. Concretely, the result showed that the GBM model had more large-scale prediction errors than the XGBoost and RF models.

To improve the interpretability of the models, the XAI methods SHAP and ICE plots were used to show the determinant factors amongst the features that influence the price prediction.

Consequently, for all the models, the SHAP plot showed that "Age, BMI, AnyChronicDiseases, and HistoryofCancerInFamily" were the most influential factors determining the premium price, while "Diabetes and knownAllergies" had negligible effects on the premium price. Also, "BloodPressureProblems" had varying effects on the premium price for each model; for the "NumberOfMajorSurgeries", a higher number negatively influenced the premium price, while a lower number increased the premium price for all the models.

Furthermore, the insights from the ICE plots showed that for all 3 models, "Age and BMI" consistently had the highest interactions and were the most influential features for price determination. Also, "AnyTransplant and AnyChronicDiseases" showed major price increases as the feature value increased for all 3 models. For XGBoost and GBM models, "HistoryOfCancerInFamily and BloodPressureProblems" showed minimal increase in price as the feature values increased but for the RF model, both features showed a higher impact on the price as the feature values increased. Additionally, for the XGBoost and GBM models, the "KnownAllergies and Diabetes" features had a negligible impact on the price determination while for the RF model, even though the "KnownAllergies" feature maintained a relatively negligible impact on the price, the "Diabetes" feature seemed to show a decrease in price as the feature values increased. Finally, for all 3 models, the "NumberOfMajorSurgeries" feature maintained a reduced price when the feature value increased and vice versa.

Both XAi methods seemed to come to the same conclusion in identifying the key determinant features that influenced the PremiumPrices. However, when comparing the outcome of both XAi methods, we see that the ICE plots showed in more detail the interactions between each variable than the SHAP analysis which seemed to be more high-level.

Concretely, this study aligns with Bora *et al*. (2022) and Langenberger *et al*. (2023) in showcasing the effectiveness of XAi methods in explaining black-box ML models for medical insurance cost predictions and identifying how the variables influenced the premium prices. From the study, it can be inferred that Age and BMI were the most prevalent features that determined medical insurance costs among the 3 models. Various studies on Actuarial Analysis have also backed our findings including Fujita *et al*. (2018), Kamble *et al.* (2018), Barnes (2015), Willie (2015), and Braver *et al.* (2004).

The significance of this study to the medical insurance industry centers on providing better decision support to potential medical insurance buyers as they are usually inundated with so many options to choose from. This study's contribution could help insurers design an insurance feature filtering function for customers based on the order of importance of the key determinants and their potential impact on the premium price. This helps improve the efficiency and transparency of the search process, thus making it easier for medical insurance buyers to select policies that best meet their specific situation and financial status. This research could also prove vital to policymakers and regulators within the health insurance industry as they seek more transparency in the use of AI technology for the Insurance underwriting process.

Since AI uses data to generate insights, it is also important to mention the challenges that come with this technology. Using AI in the medical insurance underwriting process is difficult due to limited access to data, especially for independent researchers. To provide the best recommendation on insurance and costs, one usually must know the details of someone's life, health, family, etc. This information is understandably scarce because both hospitals and insurance companies are bound by confidentiality. However, quality primary data sources would be required to get the best out of any ML project. Thus, for this study, the major limitations include;

1. Limited access to high-quality, comprehensive medical and insurance data which is crucial for accurate predictions.
2. The available dataset is limited in terms of sample size, making it challenging to build more robust ML models like deep learning models.
3. Inability to validate the performance of our ML models in real-world insurance settings due to the complexity of insurance policies and the associated red tape.

Addressing these limitations requires collaboration between researchers, insurance providers, and regulatory bodies. For future research, we recommend collaboration with insurance underwriters and sourcing more data sources (primary and secondary data) as it can improve the models' prediction accuracy and generalization.

# Appendix

Link to the programming codes for this study on Kaggle: Predicting Medical Insurance Cost using ML and XAi | Kaggle

Beyond Explainable AI. xXAi 2020. Lecture Notes in Computer Science(), vol 13200. Springer, Cham. https://doi.org/10.1007/978-3-031-04083-2_2

Ikerionwu, C., Ugwuishiwu, C., Okpala, I., James, I., Okoronkwo, M., Nnadi, C., Orji, U., Ebem, D. and Ike, A., (2022). Application of Machine and Deep Learning Algorithms in Optical Microscopic Detection of Plasmodium Parasites: A Malaria Diagnostic Tool for the Future. Photodiagnosis and Photodynamic Therapy, 103198. https://doi.org/10.1016/j.pdpdt.2022.103198

ISSA (2021). Improving health insurance systems, coverage, and service quality. [Online] https://ww1.issa.int/analysis/improving-health-insurance-systems-coverage-and-service-quality, Date Accessed: 06/03/2023.

Janizek, J. D., Celik, S., & Lee, S. I. (2018). Explainable machine learning prediction of synergistic drug combinations for precision cancer medicine. BioRxiv, 331769. https://doi.org/10.1101/331769

Jordan, S., Paul, H. L., & Philips, A. Q. (2022a). Online Appendix for How to Cautiously Uncover the 'Black Box' of Machine Learning Models for Legislative Scholars.

Jordan, S., Paul, H. L., & Philips, A. Q. (2022b). How to Cautiously Uncover the "Black Box" of Machine Learning Models for Legislative Scholars. Legislative Studies Quarterly. https://doi.org/10.1111/lsq.12378

Kagan, J. (2023). Health Insurance: Definition, How It Works. Investopedia. [Online] https://www.investopedia.com/terms/h/healthinsurance.asp. Date Accessed: 06/03/2023.

Kaggle (2021). Medical Insurance Premium Prediction. [Online] https://www.kaggle.com/datasets/tejashvi14/medical-insurance-premium-prediction. Date Accessed: 06/03/2023.

Kamble, P. S., Hayden, J., Collins, J., Harvey, R. A., Suehs, B., Renda, A., ... & Bouchard, J. (2018). Association of obesity with healthcare resource utilization and costs in a commercial population. Current medical research and opinion, 34(7), 1335-1343. https://doi.org/10.1080/03007995.2018.1464435

Kang, L., Hu, G., Huang, H., Lu, W., & Liu, L. (2020). Urban traffic travel time short-term prediction model based on spatio-temporal feature extraction. journal of advanced transportation, 2020, 1-16. https://doi.org/10.1155/2020/3247847

Kiatkarun, K., & Phunchongharn, P. (2020, September). Automatic Hyper-Parameter Tuning for Gradient Boosting Machine. In 2020 1st International Conference on Big Data Analytics and Practices (IBDAP) (pp. 1-6). IEEE. https://doi.org/10.1109/IBDAP50342.2020.9245609

Klein, E. (2012, March 2). High health-care costs: It's all in the pricing. The Washington Post. [Online] http://www.washingtonpost.com/business/high-health-care-costs-its-all-in-the-pricing/2012/02/28/gIQAtbhimR_story.html

Konstantinov, A., Utkin, L., & Muliukha, V. (2021, January). Gradient boosting machine with partially randomized decision trees. In 2021 28th Conference of Open Innovations Association (FRUCT) (pp. 167-173). IEEE. https://doi.org/10.23919/FRUCT50888.2021.9347631

Kshirsagar, R., Hsu, L.-Y., Greenberg, C. H., McClelland, M., Mohan, A., Shende, W., Tilmans, N. P., Guo, M., Chheda, A., Trotter, M., Ray, S., & Alvarado, M. (2021). Accurate and Interpretable Machine
38